\documentclass{article}

\usepackage{microtype}
\usepackage{graphicx}
\usepackage{subfigure}
\usepackage{amsmath}
\usepackage{amsfonts}
\usepackage{amssymb}
\usepackage{booktabs} % for professional tables
\usepackage{amsthm}
\usepackage[dvipsnames]{xcolor}

\usepackage[pdftitle=Learning Adversarial MDPs with Bandit Feedback and Unknown Transition,pdflang=en-US]{hyperref}

\usepackage{xspace}
\usepackage{color-edits}
\addauthor{HL}{Green}
\addauthor{CJ}{Red}
\addauthor{SS}{Blue}

\newcommand{\field}[1]{\mathbb{#1}}

\newcommand{\fR}{\field{R}}

\newcommand{\E}{\field{E}}

\newcommand{\order}{\ensuremath{\mathcal{\tilde{O}}}}

\newcommand{\Ind}[1]{ \field{I}_{\{{#1}\}} }

\newcommand{\scO}{\mathcal{O}}

\newcommand{\calP}{\mathcal{P}}
\newcommand{\calF}{\mathcal{F}}

\newcommand{\alg}{\textsc{UOB-REPS}\xspace}
\newcommand{\ind}{\mathbb{I}}
\newcommand{\qhat}{\widehat{q}}
\newcommand{\ellhat}{\widehat{\ell}}
\newcommand{\Phat}{\widehat{P}}
\newcommand{\Shat}{\widehat{S}}
\newcommand{\xtilde}{\tilde{x}}

\newcommand{\regone}{\textsc{Error}}
\newcommand{\regtwo}{\textsc{Bias}_1}
\newcommand{\regthree}{\textsc{Reg}}
\newcommand{\regfour}{\textsc{Bias}_2}

\newtheorem{theorem}{Theorem}
\newtheorem{lemma}[theorem]{Lemma}

\allowdisplaybreaks

\usepackage{amsmath}
\usepackage{mathtools}
\DeclareMathOperator*{\argmin}{\arg\!\min}
\DeclareMathOperator*{\argmax}{\arg\!\max}

\newlength\myindent
\usepackage[accepted]{icml2020}

\icmltitlerunning{Learning Adversarial MDPs with Bandit Feedback and Unknown Transition}

\begin{document}

\twocolumn[
\icmltitle{Learning Adversarial Markov Decision Processes \\ with Bandit Feedback and Unknown Transition}

% It is OKAY to include author information, even for blind
% submissions: the style file will automatically remove it for you
% unless you've provided the [accepted] option to the icml2019
% package.

% List of affiliations: The first argument should be a (short)
% identifier you will use later to specify author affiliations
% Academic affiliations should list Department, University, City, Region, Country
% Industry affiliations should list Company, City, Region, Country

% You can specify symbols, otherwise they are numbered in order.
% Ideally, you should not use this facility. Affiliations will be numbered
% in order of appearance and this is the preferred way.
%\icmlsetsymbol{equal}{*}

\begin{icmlauthorlist}
\icmlauthor{Chi Jin}{Princeton}
\icmlauthor{Tiancheng Jin}{USC}
\icmlauthor{Haipeng Luo}{USC}
\icmlauthor{Suvrit Sra}{MIT}
\icmlauthor{Tiancheng Yu}{MIT}
\end{icmlauthorlist}

\icmlaffiliation{Princeton}{Princeton University}
\icmlaffiliation{USC}{University of Southern California}
\icmlaffiliation{MIT}{Massachusetts Institute of Technology}

\icmlcorrespondingauthor{Tiancheng Jin}{tiancheng.jin@usc.edu}
\icmlcorrespondingauthor{Tiancheng Yu}{yutc@mit.edu}

% You may provide any keywords that you
% find helpful for describing your paper; these are used to populate
% the "keywords" metadata in the PDF but will not be shown in the document
%\icmlkeywords{Machine Learning, ICML}

\vskip 0.3in]

% this must go after the closing bracket ] following \twocolumn[ ...

% This command actually creates the footnote in the first column
% listing the affiliations and the copyright notice.
% The command takes one argument, which is text to display at the start of the footnote.
% The \icmlEqualContribution command is standard text for equal contribution.
% Remove it (just {}) if you do not need this facility.

\printAffiliationsAndNotice{}  % leave blank if no need to mention equal contribution
%\printAffiliationsAndNotice{\icmlEqualContribution} % otherwise use the standard text.

\begin{abstract}
We consider the task of learning in episodic finite-horizon Markov decision processes with an unknown transition function, bandit feedback, and adversarial losses. 
We propose an efficient algorithm that achieves $\mathcal{\tilde{O}}(L|X|\sqrt{|A|T})$ regret with high probability, where $L$ is the horizon, $|X|$ the number of states, $|A|$ the number of actions, and $T$ the number of episodes. 
To our knowledge, our algorithm is the first to ensure $\mathcal{\tilde{O}}(\sqrt{T})$ regret in this challenging setting; in fact it achieves the same regret as~\citep{rosenberg19a} who consider the easier setting with full-information. 
Our key contributions are two-fold: a tighter confidence set for the transition function; and an optimistic loss estimator that is inversely weighted by an {\it upper occupancy bound}.
\end{abstract}

% !TEX root = main.tex

\vspace*{-14pt}
\section{Introduction}
Reinforcement learning studies the problem where a learner interacts with the environment sequentially and aims to improve her strategy over time.
The environment dynamics are usually modeled as a Markov Decision Process (MDP) with a fixed and unknown transition function. 
We consider a general setting where the interaction proceeds in episodes with a fixed horizon. Within each episode the learner sequentially observes her current state, selects an action, suffers and observes the loss corresponding to the chosen state-action pair, and then transits to the next state according to the underlying transition function.\footnote{%
As in previous work~\citep{rosenberg19a,rosenberg2019online}, throughout we use the term ``losses'' instead of ``rewards'' to be consistent with the adversarial online learning literature. One can translate between losses and rewards by simply taking negation.
}
The goal of the learner is to minimize her regret: the difference between her total loss and the total loss of an optimal fixed policy. 

The majority of the literature in learning MDPs assumes stationary losses, that is, the losses observed for a specific state-action pair follow a fixed and unknown distribution. To better capture applications with non-stationary or even adversarial losses, the works \citep{Even-Dar:2009:OMD:1599452.1599466, yu2009markov} are among the first to study the problem of learning adversarial MDPs, where the losses can change arbitrarily between episodes. 
There are several follow-ups in this direction, such as~\citep{yu2009markov, neu2010online, neu2012unknown, zimin2013, dekel2013better, rosenberg19a}.
See Section~\ref{sec:related} for more related work.

For an MDP with $|X|$ states, $|A|$ actions, $T$ episodes, and $L$ steps in each episode, the best existing result is the work~\citep{rosenberg19a}, which achieves $\order(L|X|\sqrt{|A|T})$ regret, assuming a fixed and unknown transition function, adversarial losses, but importantly {\it full-information} feedback: i.e., the loss for {\it every} state-action pair is revealed at the end of each episode.
On the other hand, with the more natural and standard {\it bandit feedback} (where only the loss for each visited state-action pair is revealed),
a later work by the same authors~\citep{rosenberg2019online} achieves regret $\order(L^{3/2}|X||A|^{1/4}T^{3/4})$, which has a much worse dependence on the number of episodes $T$ compared to the full-information setting.

Our main contribution significantly improves on~\citep{rosenberg2019online}. In particular, we propose an efficient algorithm that achieves $\order(L|X|\sqrt{|A|T})$ regret in the same setting with bandit feedback, an unknown transition function, and adversarial losses.
Although our regret bound still exhibits a gap compared to the best existing lower bound $\Omega(L\sqrt{|X||A|T})$~\citep{jin2018q},
to the best of our knowledge, for this challenging setting our result is the first  to achieve $\order(\sqrt{T})$ regret. Importantly, this also matches the regret upper bound of~\citet{rosenberg19a}, who consider the easier setting with full-information feedback.

Our algorithm builds on the UC-O-REPS algorithm~\citep{rosenberg19a, rosenberg2019online}---we also construct confidence sets to handle the unknown transition function, and apply Online Mirror Descent over the space of occupancy measures (see Section~\ref{sec:occupancy}) to handle adversarial losses.
The first key difference and challenge is that with bandit feedback, to apply Online Mirror Descent we must construct good loss estimators since the loss function is not completely revealed. However, the most natural approach of building unbiased loss estimators via inverse probability requires knowledge of the transition function, and is thus infeasible in our setting.

We address this key challenge by proposing a novel biased and optimistic loss estimator (Section~\ref{sec:estimator}). Specifically, instead of inversely weighting the observation by the probability of visiting the corresponding state-action pair (which is unknown), we use the \emph{maximum probability among all plausible transition functions} specified by a confidence set, which we call \emph{upper occupancy bound}. This idea resembles the optimistic principle of using {\it upper confidence bounds} for many other problems of learning with bandit feedback, such as stochastic multi-armed bandits~\citep{auer2002finite}, stochastic linear bandits~\citep{chu2011contextual, abbasi2011improved}, and reinforcement learning with stochastic losses~\citep{jaksch2010near, azar2017minimax, jin2018q}.
However, as far as we know, applying optimism in constructing loss estimators for an adversarial setting is new.

The second key difference of our algorithm from UC-O-REPS (Section~\ref{sec:conf_sets}) lies in a new confidence set for the transition function. %that it uses to control the bias of the loss estimators. 
Specifically, for each state-action pair, the confidence set used in UC-O-REPS and previous works such as~\citep{jaksch2010near, azar2017minimax} imposes a total variation constraint on the transition probability, while our proposed confidence set imposes \emph{an independent constraint on the transition probability for each next state,}
% around the empirical transition, while our proposed confidence set imposes {\it an independent constraint for each next state} 
and is strictly tighter. 
Indeed, with the former we can only prove an $\order(L|X|^2\sqrt{|A|T})$ regret, while with the latter we improve it to $\order(L|X|\sqrt{|A|T})$.
Analyzing the non-trivial interplay between our optimistic loss estimators and the new confidence set is one of our key technical contributions.

Finally, we remark that our proposed upper occupancy bounds can be computed efficiently via backward dynamic programming and solving some linear programs greedily, and thus our algorithm can be implemented efficiently.

%%% Local Variables:
%%% mode: latex
%%% TeX-master: "main"
%%% End:

% !TEX root = main.tex

\subsection{Related Work}
\label{sec:related}

\paragraph{Stochastic losses.}
Learning MDPs with stochastic losses and bandit feedback is relatively well-studied for the tabular case (that is, finite number of states and actions).
For example, in the episodic setting, using our notation,\footnote{%
We warn the reader that in some of these cited papers, the notation $|X|$ or $T$ might be defined differently (often $L$ times smaller for $|X|$ and $L$ times larger for $T$).
We have translated the bounds based on Table~1 of~\citep{jin2018q} using our notation defined in Section~\ref{sec:probform}.
} 
the UCRL2 algorithm of~\citet{jaksch2010near} achieves $\order(\sqrt{L^3|X|^2|A|T})$ regret,
and the UCBVI algorithm of~\citet{azar2017minimax} achieves the optimal bound $\order(L\sqrt{|X||A|T})$, 
both of which are model-based algorithms and construct confidence sets for both the transition function and the loss function.
The recent work~\citep{jin2018q} achieves a suboptimal bound $\order(\sqrt{L^3|X||A|T})$ via an optimistic Q-learning algorithm that is model-free.
Besides the episodic setting, other setups such as discounted losses or infinite-horizon average-loss setting have also been heavily studied; see for example~\citep{ouyang2017learning, fruit2018efficient, zhang2019regret, wei2019model, dong2019q} for some recent works.

\vskip3pt
\noindent\textbf{Adversarial losses.}
Based on whether the transition function is known and whether the feedback is full-information or bandit, we discuss four categories separately.

{\it Known transition and full-information feedback.}
Early works on adversarial MDPs assume a known transition function and full-information feedback.
For example, \citet{Even-Dar:2009:OMD:1599452.1599466} propose the algorithm MDP-E and prove a regret bound of $\order(\tau^2\sqrt{T\ln|A|})$ where $\tau$ is the mixing time of the MDP; another work~\citep{yu2009markov} achieves $\order(T^{2/3})$ regret. Both of these consider a continuous setting (as opposed to the episodic setting that we study).
Later~\citet{zimin2013} consider the episodic setting and propose the O-REPS algorithm which applies Online Mirror Descent over the space of occupancy measures, a key component adopted by~\citep{rosenberg19a} and our work.
O-REPS achieves the optimal regret $\order(L\sqrt{T\ln(|X||A|)})$ in this setting.
 
{\it Known transition and bandit feedback.}
Several works consider the harder bandit feedback model while still assuming known transitions. The work~\citep{neu2010online} achieves regret $\order(L^2\sqrt{T|A|}/\alpha)$, assuming that all states are reachable with some probability $\alpha$ under all policies.
Later, \citet{neu2014online} eliminates the dependence on $\alpha$ but only achieves $\order(T^{2/3})$ regret. 
The O-REPS algorithm of~\citep{zimin2013} again achieves the optimal regret $\order(\sqrt{L|X||A|T})$.
Another line of works~\citep{arora2012deterministic, dekel2013better} assumes deterministic transitions for a continuous setting without some unichain structure,
which is known to be harder and suffers $\Omega(T^{2/3})$ regret~\citep{dekel2014bandits}.

{\it Unknown transition and full-information feedback.}
To deal with unknown transitions, \citet{neu2012unknown} propose the Follow the Perturbed Optimistic Policy algorithm and achieve $\order(L|X||A|\sqrt{T})$ regret.
Combining the idea of confidence sets and Online Mirror Descent, the UC-O-REPS algorithm of~\citep{rosenberg19a} improves the regret to $\order(L|X|\sqrt{|A|T})$.
We note that this work also studies general convex performance criteria, which we do not consider.

{\it Unknown transition and bandit feedback.}
This is the setting considered in our work.
The only previous work we are aware of~\citep{rosenberg2019online} achieves a regret bound of $\order(T^{3/4})$, or $\order(\sqrt{T}/\alpha)$ under the strong assumption that under any policy,  all states are reachable with probability $\alpha$ that could be arbitrarily small in general.
Our algorithm achieves $\order(\sqrt{T})$ regret without this assumption by using a different loss estimator and by using a tighter confidence set.
We also note that the lower bound of $\Omega(L\sqrt{|X||A|T})$~\citep{jin2018q} still applies.

\paragraph{Adversarial transition functions.}
There exist a few works that consider both time-varying transition functions and time-varying losses~\citep{yu2009arbitrarily, cheung2019reinforcement, lykouris2019corruption}.
Most recently, \citet{lykouris2019corruption} consider a stochastic problem with $C$ episodes arbitrarily corrupted and obtain $\order(C\sqrt{T}+C^2)$ regret (ignoring dependence on other parameters). This bound is of order $\order(\sqrt{T})$ only when $C$ is a constant, and is vacuous whenever $C = \Omega(\sqrt{T})$. In comparison, our bound is always $\order(\sqrt{T})$ no matter how much corruption there is in the losses, but our algorithm cannot handle changing transition functions.

%The rest paper is organized as follows. In Section~\ref{sec:probform}, we introduce the problem formulation and setting, including the assumptions and notations. We present our algorithm and its three key components in Section~\ref{sec:alg}, discuss how these components help address the difficulties of learning in this setting. Finally, we give a proof sketch of the achieved regret bound in Section~\ref{sec:analysis}.

%%% Local Variables:
%%% mode: latex
%%% TeX-master: "main"
%%% End:

% !TEX root = main.tex

\section{Problem Formulation}
\label{sec:probform}
An adversarial Markov decision process is defined by a tuple $(X,A,P, \{\ell_t\}_{t=1}^T)$, 
where $X$ is the finite state space, $A$ is the finite action space, $P:X \times A \times X \rightarrow [0,1]$ is the transition function, with $P(x'|x,a)$ being the probability of transferring to state $x'$ when executing action $a$ in state $x$,
and $\ell_t: X\times A\rightarrow [0,1]$ is the loss function for episode $t$.
%and $\mathcal{L} = \{\ell | \ell: X \times A \rightarrow [0,1]\}$ is the function class of all possible loss function $\ell$.

%Each entry $P(x'|x,a)$ of transition function $P$, is the probability that the next state yielded by Markovian environment will be $x'$, given that action $a$ is selected at the state $x$. 

In this work, we consider an episodic setting with finite horizons and assume that the MDP has a layered structure, satisfying the following conditions: 
\begin{itemize}
    \item The state space $X$ consists of $L+1$ layers $X_0, \ldots, X_L$ such that $X$ = $\bigcup_{k=0}^{L}X_k$ and $X_i \cap X_j = \emptyset$ for $i\neq j$.
    \item $X_0$ and $X_L$ are singletons, that is, $X_0 = \{x_0\}$ and $X_L = \{x_L\}$.
    \item Transitions are possible only between consecutive layers. In other words, if $P(x'|x,a) > 0$, then $x' \in X_{k+1}$ and $x\in X_k$ for some $k$. 
\end{itemize}

These assumptions were made in previous work~\citep{neu2012unknown, zimin2013, rosenberg19a} as well.
They are not necessary but greatly simplify notation and analysis.
Such a setup is sometimes referred to as the loop-free stochastic shortest path problem in the literature.
It is clear that this is a strict generalization of the episodic setting studied in~\citep{azar2017minimax, jin2018q} for example, where the number of states is the same for each layer (except for the first and the last one).\footnote{%
In addition, some of these works (such as~\citep{azar2017minimax}) also assume that the states have the same name for different layers, and the transition between the layers remains the same.
Our setup does not make this assumption and is closer to that of~\citep{jin2018q}.
We also refer the reader to footnote~2 of~\citep{jin2018q} for how to translate regret bounds between settings with and without this extra assumption.
}
We also point out that our algorithms and results can be easily modified to deal with
% results directly apply to 
a more general setup where the first layer has multiple states and in each episode the initial state is decided adversarially, as in~\citep{jin2018q} (details omitted).

%These assumptions are first proposed in  \citet{neu2010online} and soon adopted by \citet{zimin2013}, \citet{rosenberg19a} et cetera. It's a strict generalization of the episodic setting introduced in \citet{Sutton1998}. Besides there are other generalizations such as in \citet{azar2017minimax} and \citet{jin2018q}. 

% \CJcomment{Protocol 1 looks like we can only deal with oblivious adversarial loss, while our proof should apply to adaptive adversary. I think we should define MDP as (X, A, P) only, and let adversary to pick a possibly adaptive loss l\_t in the beginning of episode l\_t.}

The interaction between the learner and the environment is presented in Protocol~\ref{alg:LEInteract}. Ahead of time, the environment decides an MDP, and only the state space $X$ with its layer structure and the action space $A$ are known to the learner.
% In particular, the loss functions $\ell_1, \ldots, \ell_T$ can be chosen adversarially with the knowledge of the learner's algorithm.
%
The interaction proceeds in $T$ episodes. In episode $t$, the adversary decides the loss function $\ell_t$, which can depend on the learner's algorithm and the randomness before episode $t$. Simultaneously, the learner starts from state $x_0$ and decides a stochastic policy $\pi_t:X\times A \rightarrow [0,1]$, where $\pi_t(a|x)$ is the probability of taking action $a$ at a given state $x$, so that $\sum_{a\in A}\pi_t(a|x)=1$ for every state $x$. Then, the learner executes this policy in the MDP, generating $L$ state-action pairs $(x_0, a_0), \ldots, (x_{L-1}, a_{L-1})$.\footnote{%
Formally, the notation $(x_0, a_0), \ldots, (x_{L-1}, a_{L-1})$ should have a $t$ dependence. Throughout the paper we omit this dependence for conciseness as it is clear from the context.
}
Specifically, for each $k=0, \ldots, L-1$, action $a_k$ is drawn from $\pi_t(\cdot|x_k)$ and the next state $x_{k+1}$ is drawn from $P(\cdot|x_k, a_k)$.

Importantly, instead of observing the loss function $\ell_t$ at the end of episode $t$ \cite{rosenberg19a}, in our setting the learner only observes the loss for each visited state-action pair: $\ell_t(x_0, a_0), \ldots, \ell_t(x_{L-1}, a_{L-1})$.
That is, we consider the more challenging setting with bandit feedback.

\makeatletter
\renewcommand{\ALG@name}{Protocol}
\makeatother

\begin{algorithm}[tb]
   \caption{Learner-Environment Interaction}
   \label{alg:LEInteract}
\begin{algorithmic}
   \STATE {\bfseries Parameters:} state space $X$ and action space $A$ (known to the learner), unknown transition function $P$
   \FOR{$t=1$ {\bfseries to} $T$}
   \STATE adversary decides a loss function $\ell_t: X\times A \rightarrow [0,1]$
   \STATE learner decides a policy $\pi_t$ and starts in state $x_0$
   \FOR{$k=0$ {\bfseries to} $L-1$}
   \STATE learner selects action $a_k \sim \pi_t(\cdot|x_k)$
   \STATE learner observes loss $\ell_t(x_k,a_k)$
   \STATE environment draws a new state $x_{k+1} \sim P(\cdot|x_k,a_k)$
   \STATE learner observes state $x_{k+1}$
   \ENDFOR   
   \ENDFOR
\end{algorithmic}
\end{algorithm}

\makeatletter
\renewcommand{\ALG@name}{Algorithm}
\makeatother

For any given policy $\pi$, we denote its expected loss in episode $t$ by
\[
\mathbb{E}\left[\left.\sum_{k=0}^{L-1}\ell_t(x_k,a_k)\right| P,\pi\right],
\]
where the notation $\mathbb{E}[\cdot|P,\pi]$ emphasizes that the state-action pairs $(x_0, a_0), \ldots, (x_{L-1}, a_{L-1})$ are random variables generated according to the transition function $P$ and a stochastic policy $\pi$. 
The total loss over $T$ episodes for any fixed policy $\pi$ is thus
\[
L_T(\pi) = \sum_{t=1}^{T}\mathbb{E}\left[\left.\sum_{k=0}^{L-1}\ell_t(x_k,a_k)\right| P,\pi\right],
\]
while the total loss of the learner is
\begin{equation*}
\begin{split}
L_T & = \sum_{t=1}^{T}\mathbb{E}\left[\left.\sum_{k=0}^{L-1}\ell_t(x_k,a_k)\right| P,\pi_t\right].
\end{split}
\end{equation*}
The goal of the learner is to minimize the regret, defined as
\[
R_T = L_T - \min_{\pi}L_T(\pi)
\]
where $\pi$ ranges over all stochastic policies. 

\paragraph{Notation.}
We use $k(x)$ to denote the index of the layer to which state $x$ belongs, and $\ind\{\cdot\}$ to denote the indicator function whose value is $1$ if the input holds true and $0$ otherwise.
Let $o_t = \{(x_k, a_k, \ell_t(x_k, a_k))\}_{k=0}^{L-1}$ be the observation of the learner in episode $t$,
and $\calF_t$ be the $\sigma$-algebra generated by $(o_1, \ldots, o_{t-1})$.
Also let $\E_t[\cdot]$ be a shorthand of $\E[\cdot | \calF_{t}]$.

\subsection{Occupancy Measures}
\label{sec:occupancy}

Solving the problem with techniques from online learning requires introducing the concept of \textit{occupancy measures}~\citep{altman1999constrained,neu2012unknown}.
Specifically, the occupancy measure $q^{P,\pi}: X \times A \times X \rightarrow [0,1]$ associated with a stochastic policy $\pi$ and a transition function $P$ is defined as
\[
q^{P,\pi}(x,a,x') = \Pr\left[x_k=x,a_k=a,x_{k+1}=x' \;|\; P,\pi\right],
\]
where $k=k(x)$ is the index of the layer to which $x$ belongs.
In other words, $q^{P,\pi}(x,a,x')$ is the marginal probability of encountering the triple $(x, a, x')$ when executing policy $\pi$ in a MDP with transition function $P$.

Clearly, an occupancy measure $q$ satisfies the following two properties.
First, due to the loop-free structure, each layer is visited exactly once and thus for every $k = 0, \ldots, L-1$,
\begin{equation}
\sum_{x\in X_k}\sum_{a\in A}\sum_{x'\in X_{k+1}}q(x,a,x') = 1. 
\label{eq:1}
\end{equation}
Second, the probability of entering a state when coming from the previous layer is exactly the probability of leaving from that state to the next layer (except for $x_0$ and $x_L$). Therefore, for every $k = 1, \ldots, L-1$ and every state $x\in X_k$, we have 
\begin{equation}
\sum_{x'\in X_{k-1}}\sum_{a\in A}q(x',a,x) = \sum_{x'\in X_{k+1}}\sum_{a\in A}q(x,a,x').
\label{eq:2}
\end{equation}

It turns out that these two properties suffice for any function $q: X\times A\times A \rightarrow [0,1]$ to be an occupancy measure associated with some transition function and some policy.

\begin{lemma}[\citet{rosenberg19a}]
\label{lem:occupancy_measure}
If a function $q: X \times A \times X \rightarrow[0,1]$ satisfies conditions~\eqref{eq:1} and~\eqref{eq:2}, then it is a valid occupancy measure associated with the following induced transition function $P^q$ and induced policy $\pi^q$:
\begin{equation*}
\begin{split}
P^q(x'|x,a) & = \frac{q(x,a,x')}{\sum_{y\in X_{k(x)+1}}q(x,a,y)}, \\ 
\pi^q(a|x) & = \frac{\sum_{x'\in X_{k(x)+1}}q(x,a,x')}{\sum_{b\in A}\sum_{x'\in X_{k(x)+1}}q(x,b,x')}. 
\end{split}
\end{equation*}
\end{lemma}

We denote by $\Delta$ the set of valid occupancy measures, that is, the subset of $[0,1]^{X \times A \times X}$ satisfying conditions~\eqref{eq:1} and~\eqref{eq:2}. 
For a fixed transition function $P$, we denote by $\Delta(P) \subset \Delta$ the set of occupancy measures whose induced transition function $P^q$ is exactly $P$. 
Similarly, we denote by $\Delta(\calP)\subset\Delta$ the set of occupancy measures whose induced transition function $P^q$ belongs to a set of transition functions $\calP$. 

With the concept of occupancy measure, we can reduce the problem of learning a policy to the problem of learning an occupancy measure and apply online linear optimization techniques. Specifically, with slight abuse of notation, for an occupancy measure $q$ we define
\[
q(x,a) = \sum_{x' \in X_{k(x)+1}}q(x,a,x')
\]
for all $x \neq x_L$ and $a\in A$, which is the probability of visiting state-action pair $(x,a)$.
Then the expected loss of following a policy $\pi$ for episode $t$ can be rewritten as
\begin{equation*}
\begin{split}
&\mathbb{E}\left[\left.\sum_{k=0}^{L-1}\ell_t(x_k,a_k)\right|P,\pi\right] \\
= & \sum_{k=0}^{L-1}\sum_{x\in X_k}\sum_{a\in A} q^{P,\pi}(x,a) \ell_t(x,a) \\
=  &\sum_{x\in X\setminus \{x_L\}, a\in A}q^{P,\pi}(x,a)\ell_t(x,a) \triangleq  \langle q^{P,\pi}, \ell_t\rangle,
\end{split}
\end{equation*}
and accordingly the regret of the learner can be rewritten as
\begin{equation}
\label{eq:regdef}
R_T = L_T - \min_{\pi}L_T(\pi) = \sum_{t=1}^T \langle q^{P,\pi_t} - q^*, \ell_t \rangle,
\end{equation}
where $q^* \in \argmin_{q\in\Delta(P)} \sum_{t=1}^T\langle q,\ell_t\rangle$ is the optimal occupancy measure in $\Delta(P)$.

On the other hand, assume for a moment that the set $\Delta(P)$ were known and the loss function $\ell_t$ was revealed at the end of episode $t$.
Consider an online linear optimization problem (see~\citep{hazan2016introduction} for example) with decision set $\Delta(P)$ and linear loss parameterized by $\ell_t$ at time $t$. 
In other words, at each time $t$, the learner proposes $q_t \in \Delta(P)$ and suffers loss $\langle q_t, \ell_t \rangle$.
The regret of this problem is
\begin{equation}
\label{eq:ocponline}
\sum_{t=1}^T \langle q_t - q^*, \ell_t \rangle.    
\end{equation}
Therefore, if in the original problem, we set $\pi_t = \pi^{q_t}$, then the two regret measures Eq.~\eqref{eq:regdef} and Eq.~\eqref{eq:ocponline} are exactly the same by Lemma~\ref{lem:occupancy_measure} and we have thus reduced the problem to an instance of online linear optimization.

It remains to address the issues that $\Delta(P)$ is unknown and that we have only partial information on $\ell_t$. The first issue can be addressed by constructing a confidence set $\calP$ based on observations and replacing $\Delta(P)$ with $\Delta(\calP)$, 
and the second issue is addressed by constructing loss estimators with reasonably small bias and variance.
For both issues, we propose new solutions compared to~\citep{rosenberg2019online}.

Note that importantly, the above reduction does not reduce the problem to an instance of the well-studied bandit linear optimization~\citep{abernethy2008competing} where the quantity $\langle q_t, \ell_t \rangle$ (or a sample with this mean) is observed.
Indeed, roughly speaking, what we observed in our setting are samples with mean $\langle q^{P, \pi^{q_t}}, \ell_t \rangle$.
These two are different when we do not know $P$ and have to operate over the set $\Delta(\calP)$.

%%% Local Variables:
%%% mode: latex
%%% TeX-master: "main"
%%% End:

% !TEX root = main.tex

\section{Algorithm}
\label{sec:alg}
The complete pseudocode of our algorithm, \alg, is presented in Algorithm~\ref{alg:main}.
The three key components of our algorithm are: 1) maintaining a confidence set of the transition function, 2) using Online Mirror Descent to update the occupancy measure, and 3) constructing loss estimators, each described in detail below.

\subsection{Confidence Sets}\label{sec:conf_sets}
The idea of maintaining a confidence set of the transition function $P$ dates back to~\citep{burnetas1997optimal}.
Specifically, the algorithm maintains counters to record the number of visits of each state-action pair $(x,a)$ and each state-action-state triple $(x,a,x')$.
To reduce the computational complexity,
a doubling epoch schedule is deployed, so that a new epoch starts whenever there exists a state-action whose counter is doubled compared to its initial value at the beginning of the epoch.
For epoch $i>1$, let $N_i(x,a)$ and $M_i(x' |x,a)$ be the initial values of the counters, that is, the total number of visits of pair $(x,a)$ and triple $(x,a,x')$ before epoch $i$.
Then the empirical transition function for this epoch is defined as
\[
\bar{P}_i(x'|x,a) = \frac{M_i(x'|x,a)}{\max\{1,N_i(x,a)\}}.
\]
Most previous works (such as~\citep{jaksch2010near, azar2017minimax, rosenberg2019online}) construct a confidence set which includes all transition functions with bounded total variation compared to $\bar{P}_i(\cdot|x,a)$ for each $(x,a)$ pair.
However, to ensure lower bias for our loss estimators, we propose a tighter confidence set which includes all transition functions with bounded distance compared to $\bar{P}_i(x'|x,a)$ for {\it each triple} $(x,a,x')$.
More specifically, the confidence set for epoch $i$ is defined as\footnote{%
It is understood that in the definition of the confidence set (Eq.~\eqref{eqn:confidence_set}), there is also an implicit constraint on $\Phat(\cdot|x,a)$ being a valid distribution over the states in $X_{k(x)+1}$, for each $(x,a)$ pair.
This is omitted for conciseness.
}
\begin{equation}\label{eqn:confidence_set}
\begin{split}
\calP_i= \Big\{\Phat: \left|\Phat(x'|x,a) - \bar{P}_{i}(x'|x,a)\right| \leq \epsilon_{i}(x'|x,a), \\
 \;\forall(x,a,x')\in X_k \times A \times X_{k+1}, k=0, \ldots, L-1 \Big\},
\end{split}
\end{equation}
where the confidence width $\epsilon_i(x'|x,a)$ is defined as
\begin{equation}\label{eqn:width}
2\sqrt{\frac{\bar{P}_{i}(x'|x,a)\ln\left(\frac{T|X||A|}{\delta}\right)}{\max\{1,N_i(x,a)-1\}}} + \frac{14\ln\left(\frac{T|X||A|}{\delta}\right)}{3\max\{1,N_i(x,a)-1\}}
\end{equation}
for some confidence parameter $\delta \in (0,1)$.
For the first epoch ($i=1$), $\calP_i$ is simply the set of all transition functions so that $\Delta(\calP_i) = \Delta$.\footnote{%
To represent $\calP_1$ in the form of Eq.~\eqref{eqn:confidence_set}, one can simply let $\bar{P}_1(\cdot|x,a)$ be any distribution and $\epsilon_1(x'|x,a) = 1$.
}

By the empirical Bernstein inequality and union bounds, one can show the following (see Appendix~\ref{app:concentration} for the proof):

\begin{lemma}\label{lem:confidence_sets}
With probability at least $1-4\delta$, we have $P\in \calP_i$ for all $i$.
\end{lemma}

Moreover, ignoring constants one can further show that our confidence bound is strictly tighter than those used in~\citep{rosenberg19a, rosenberg2019online}, which is important for getting our final regret bound (more discussions to follow in Section~\ref{sec:analysis}).

\begin{algorithm}[!htbp]
\caption{Upper Occupancy Bound Relative Entropy Policy Search (\alg)}
\label{alg:main}
\begin{algorithmic}
\STATE {\bfseries Input:} state space $X$, action space $A$, episode number $T$, learning rate $\eta$, exploration parameter $\gamma$, and confidence parameter $\delta$

\item[]
\STATE {\bfseries Initialization:}

\STATE Initialize epoch index $i=1$ and confidence set $\calP_1$ as the set of all transition functions.
\STATE For all $k=0, \ldots, L-1$ and all $(x,a,x')\in X_k\times A\times X_{k+1}$,
initialize counters
\[
N_0(x,a)=N_1(x,a)= M_0(x'|x,a)=M_1(x'|x,a)=0
\]
\STATE and occupancy measure
\[
\qhat_1(x,a,x') = \frac{1}{|X_k||A||X_{k+1}|}.
\]

\STATE  Initialize policy $\pi_1 = \pi^{\qhat_1}$.

\item[]
\FOR{ $t = 1 \ \textbf{to} \ T$ }

\STATE Execute policy $\pi_t$ for $L$ steps and obtain trajectory $x_k, a_k, \ell_t(x_k, a_k)$ for $k = 0, \ldots, L-1$.

\STATE Compute upper occupancy bound for each $k$:
\[
u_t(x_k,a_k) = \textsc{Comp-UOB}(\pi_t, x_k, a_k, \calP_{i}).
\]

\STATE Construct loss estimators for all $(x,a)$:
\[
\ellhat_t(x,a) = \frac{\ell_t(x, a)}{u_t(x,a)+\gamma}\ind\{x_{k(x)} = x, a_{k(x)} = a\}.
\]

\STATE Update counters: for each $k$,
\begin{align*}
N_i(x_k, a_k) &\gets N_i(x_k, a_k) + 1, \\
M_i(x_{k+1} | x_k, a_k) &\gets M_i(x_{k+1} | x_k, a_k)  + 1.
\end{align*}

\IF {$\exists k, \  N_i(x_k, a_k) \geq \max\{1, 2N_{i-1}(x_k,a_k)\}$}

\STATE Increase epoch index $i \gets i+1$.

\STATE Initialize new counters: for all $(x,a,x')$,
\[
N_{i}(x,a) = N_{i-1}(x,a), M_{i}(x' | x, a)  = M_{i-1}(x' | x, a).
\]

\STATE Update confidence set $\calP_i$ based on Eq.~\eqref{eqn:confidence_set}.
\ENDIF

\item[]
\STATE Update occupancy measure ($D$ defined in Eq.~\eqref{eqn:KL}):
\[
 \qhat_{t+1} = \argmin_{q\in\Delta(\calP_i)}\; \eta \langle q, \ellhat_t \rangle + D(q \;\|\; \qhat_{t}).
\]

\STATE Update policy $\pi_{t+1} = \pi^{\qhat_{t+1}}$.
\ENDFOR

\end{algorithmic}
\end{algorithm}

\begin{algorithm}[!htbp]
\caption{\textsc{Comp-UOB}}
\label{alg:UOB}
\begin{algorithmic}
\STATE {\bfseries Input:} a policy $\pi_t$, a state-action pair $(x,a)$ and a confidence set $\calP$ of the form
\begin{small}
\[
\left\{\Phat: \left|\Phat(x'|x,a) - \bar{P}(x'|x,a)\right|\leq \epsilon(x'|x,a), \;\forall(x,a,x'
) \right\}
\]
\end{small}

\STATE {\bfseries Initialize:} for all $\xtilde \in X_{k(x)}$, set $f(\xtilde) = \ind\{\xtilde=x\}$.

\item[]
\FOR{ $k = k(x)-1 \ \textbf{to} \ 0$}
%\STATE Sort $F = \{f(x')\}_{x'\in X_{k+1}}$

\FOR{ all $\xtilde \in X_k$}

\STATE Compute $f(\xtilde)$ based on Eq.~\eqref{eqn:DP}:
\begin{small}
\begin{align*}
f(\xtilde) = \sum_{a\in A} \pi_t(a|\xtilde) \cdot \text{\textsc{Greedy}}\left(f,
\bar{P}(\cdot|\xtilde,a), \epsilon(\cdot|\xtilde,a)\right)
\end{align*}
\end{small}
(see Appendix~\ref{app:alg} for the procedure \textsc{Greedy}).

\ENDFOR
\ENDFOR

\STATE {\bfseries Return:} $\pi_t(a|x) f(x_0)$.
\end{algorithmic}
\end{algorithm}

\subsection{Online Mirror Descent (OMD)}\label{sec:OMD}
The OMD component of our algorithm is the same as~\citep{rosenberg2019online}.
As discussed in Section~\ref{sec:occupancy}, our problem is closely related to an online linear optimization problem over some occupancy measure space.
In particular, our algorithm maintains an occupancy measure $\qhat_t$ for episode $t$ and executes the induced policy $\pi_t = \pi^{\qhat_t}$.
We apply Online Mirror Descent, a standard algorithmic framework to tackle online learning problems, to update the occupancy measure as
\begin{equation}\label{eqn:update_q}
\qhat_{t+1} = \argmin_{q\in\Delta(\calP_i)}\; \eta \langle q, \ellhat_t \rangle + D(q \;\|\; \qhat_{t})
\end{equation}
where $i$ is the index of the epoch to which episode $t+1$ belongs, $\eta > 0$ is some learning rate, $\ellhat_t$ is some loss estimator for $\ell_t$, and $D(\cdot \|\cdot)$ is a Bregman divergence.
Following~\citep{rosenberg19a, rosenberg2019online}, we use the unnormalized KL-divergence as the Bregman divergence:
\begin{equation}\label{eqn:KL}
\begin{split}
D(q \;\|\; q') = & \sum_{x,a,x'}q(x,a,x')\ln\frac{q(x,a,x')}{q'(x,a,x')}  \\
&- \sum_{x,a,x'}\left(q(x,a,x')-q'(x,a,x')\right).
\end{split}
\end{equation}
Note that as pointed out earlier, ideally one would use $\Delta(P)$ as the constraint set in the OMD update, but since $P$ is unknown, using $\Delta(\calP_i)$ in place of it is a natural idea.
Also note that the update can be implemented efficiently, similarly to~\citet{rosenberg19a} (see Appendix~\ref{app:proj} for details).

\subsection{Loss Estimators}
\label{sec:estimator}
A common technique to deal with partial information in adversarial online learning problems (such as adversarial multi-armed bandits~\citep{auer2002nonstochastic}) is to construct loss estimators based on observations.
In particular, inverse importance-weighted estimators are widely applicable.
For our problem, with a trajectory $x_0, a_0, \ldots, x_{L-1}, a_{L-1}$ for episode $t$,
a common importance-weighted estimator for $\ell_t(x,a)$ would be
\[
\frac{\ell_t(x,a)}{q^{P, \pi_t}(x,a)}\ind\left\{x_{k(x)} = x, a_{k(x)} = a \right\}.
\]
Clearly this is an unbiased estimator for $\ell_t(x,a)$.
Indeed, the conditional expectation $\E_t[\ind\left\{x_{k(x)} = x, a_{k(x)} = a \right\}]$ is exactly $q^{P, \pi_t}(x,a)$ since the latter is exactly the probability of visiting $(x,a)$ when executing policy $\pi_t$ in a MDP with transition function $P$.

The issue of this standard estimator is that we cannot compute $q^{P, \pi_t}(x,a)$ since $P$ is unknown.
To address this issue, \citet{rosenberg2019online} directly use $\qhat_t(x,a)$ in place of $q^{P, \pi_t}(x,a)$, leading to an estimator that could be either an overestimate or an underestimate, and they can only show $\order(T^{3/4})$ regret with this approach.

Instead, since we have a confidence set $\calP_i$ that contains $P$ with high probability (where $i$ is the index of the epoch to which $t$ belongs),
we propose to replace $q^{P, \pi_t}(x,a)$ with an {\it upper occupancy bound}
defined as
\[
u_t(x,a) = \max_{\Phat \in \calP_i} q^{\Phat, \pi_t}(x,a),
\]
that is, the largest possible probability of visiting $(x,a)$ among all the plausible environments.
In addition, we also adopt the idea of {\it implicit exploration} from~\citep{neu2015explore} to further increase the denominator by some fixed amount $\gamma >0$.
Our final estimator for $\ell_t(x,a)$ is
\[
\ellhat_t(x,a) = \frac{\ell_t(x,a)}{u_t(x,a)+\gamma}\ind\left\{x_{k(x)} = x, a_{k(x)} = a \right\}.
\]
The implicit exploration is important for several technical reasons such as obtaining a high probability regret bound, the key motivation of the work~\citep{neu2015explore} for multi-armed bandits.
%\footnote{%
%Even if we were only aiming for an expected regret bound, we were not able to do so with $\gamma = 0$ though for technical reasons. \HLcomment{Not quite sure if we should say something like this}
%}

Clearly, $\ellhat_t(x,a)$ is a biased estimator and in particular is underestimating $\ell_t(x,a)$ with high probability (since by definition $q^{P, \pi_t}(x,a) \leq u_t(x,a)$ if $P\in\calP_i$).
The idea of using underestimates for adversarial learning with bandit feedback can be seen as an optimism principle which encourages exploration, and appears in previous work such as~\citep{AllenbergAuGyOt06, neu2015explore} in different forms and for different purposes.
A key part of our analysis is to show that the bias introduced by these estimators is reasonably small, which eventually leads to a better regret bound compared to~\citep{rosenberg2019online}.

\paragraph{Computing upper occupancy bound efficiently.}
It remains to discuss how to compute $u_t(x,a)$ efficiently.
First note that
\begin{equation}\label{eqn:UOB}
u_t(x,a) = \pi_t(a|x) \max_{\Phat \in \calP_i} q^{\Phat, \pi_t}(x)
\end{equation}
where once again we slightly abuse the notation and define $q(x) = \sum_{a'\in A} q(x,a')$ for any occupancy measure $q$, which is the marginal probability of visiting state $x$ under the associated policy and transition function.
Further define
\begin{equation*}
f(\xtilde) = \max_{\Phat \in \calP_i} \Pr\left[x_{k(x)} = x \;\big\vert\; x_{k(\xtilde)} = \xtilde, \Phat, \pi_t  \right],
\end{equation*}
for any $\xtilde$ with $k(\xtilde) \leq k(x)$,
which is the maximum probability of visiting $x$ starting from state $\xtilde$, under policy $\pi_t$ and among all plausible transition functions in $\calP_i$.
Clearly one has $u_t(x,a) = \pi_t(a|x) f(x_0)$, and also $f(\xtilde) =  \ind\{\xtilde=x\}$ for all $\xtilde$ in the same layer as $x$.
Moreover, since the confidence set $\calP_i$ imposes an independent constraint on $\Phat(\cdot|x,a)$ for each different pair $(x,a)$, we have the following recursive relation:
\begin{equation}\label{eqn:DP}
f(\xtilde) = \sum_{a\in A}\pi_t(a|\xtilde)\left( \max_{\Phat(\cdot|\xtilde,a)} \sum_{x' \in X_{k(\xtilde)+1}} \Phat(x' | \xtilde, a) f(x') \right)
\end{equation}
where the maximization is over the constraint that $\Phat(\cdot|\xtilde,a)$ is a valid distribution over $X_{k(\xtilde)+1}$ and also
\[
\left|\Phat(x'|\xtilde,a) - \bar{P}_i(x'|\xtilde,a)\right| \leq \epsilon_i(x'|\xtilde,a), \forall x'\in X_{k(\xtilde)+1}.
\]
This optimization can be solved efficiently via a greedy approach after sorting the values of $f(x')$ for all $x'\in X_{k(\xtilde)+1}$ (see Appendix~\ref{app:alg} for details).
This suggests computing $u_t(x,a)$ via backward dynamic programming from layer $k(x)$ down to layer $0$, detailed in Algorithm~\ref{alg:UOB}.

\section{Analysis}
\label{sec:analysis}

In this section, we analyze the regret of our algorithm and prove the following theorem.

\begin{theorem}\label{thm:main}
With probability at least $1-9\delta$, \alg with $\eta=\gamma=\sqrt{\frac{L\ln(L|X||A|/\delta)}{T|X||A|}}$ ensures:
\[
R_T = \scO\left(L|X|\sqrt{|A|T\ln\left(\frac{T|X||A|}{\delta}\right)}\right).
\]
\end{theorem}

The proof starts with decomposing the regret into four different terms.
Specifically, by Eq.~\eqref{eq:regdef} the regret can be written as
$R_T = \sum_{t=1}^T \langle q_t - q^*, \ell_t \rangle $ where we define $q_t = q^{P,\pi_t}$ and $q^* \in \argmin_{q\in\Delta(P)} \sum_{t=1}^T\langle q,\ell_t\rangle$.
We then add and subtract three terms and decompose the regret as
\begin{equation*}
\begin{split}
R_T &= \underbrace{\sum_{t=1}^T \left\langle q_t - \qhat_t, \ell_t \right\rangle}_{\regone}
+ \underbrace{\sum_{t=1}^T \left\langle \qhat_t , \ell_t - \ellhat_t \right\rangle}_{\regtwo} \\
&+ \underbrace{\sum_{t=1}^T \left\langle \qhat_t - q^*, \ellhat_t \right\rangle}_{\regthree}
+ \underbrace{\sum_{t=1}^T \left\langle q^*, \ellhat_t - \ell_t \right\rangle}_{\regfour}.
\end{split}
\end{equation*}

Here, the first term $\regone$ measures the error of using $\qhat_t$ to approximate $q_t$;
the third term $\regthree$ is the regret of the corresponding online linear optimization problem and is controlled by OMD;
the second and the fourth terms $\regtwo$ and $\regfour$ correspond to the bias of the loss estimators.

We bound $\regone$ and $\regtwo$ in the rest of this section. Bounding $\regthree$ and $\regfour$ is relatively standard and we defer the proofs to Appendix~\ref{app:reg_three_four}.
Combining all the bounds (specifically, Lemmas~\ref{lem:error},~\ref{lem:bias_1},~\ref{lem:OMD}, and~\ref{lem:bias_2}), applying a union bound, and plugging in the (optimal) values of $\eta$ and $\gamma$ prove Theorem~\ref{thm:main}.

Throughout the analysis we use $i_t$ to denote the index of the epoch to which episode $t$ belongs. Note that $\calP_{i_t}$ and $\qhat_t$ are both $\calF_{t}$-measurable.
We start by stating a key technical lemma which essentially describes how our new confidence set shrinks over time and is critical for bounding $\regone$ and $\regtwo$ (see Appendix~\ref{app:key_lemma} for the proof).
\begin{lemma}\label{lem:key_lemma}
With probability at least $1-6\delta$, for any collection of transition functions $\{P_t^x\}_{x\in X}$ such that $P_t^x \in \calP_{i_t}$ for all $x$, we have
\begin{equation*}
\begin{split}
 &\sum_{t=1}^{T}\sum_{x\in X, a\in A} |q^{P_t^x,\pi_t}(x,a) - q_t(x,a)| \\
 &= \scO\left(L|X|\sqrt{|A|T\ln\left(\frac{T|X||A|}{\delta}\right)}\right).
\end{split}
\end{equation*}
\end{lemma}

\paragraph{Bounding $\regone$.}
With the help of Lemma~\ref{lem:key_lemma}, we immediately obtain the following bound on $\regone$.
\begin{lemma}\label{lem:error}
With probability at least $1-6\delta$, \alg ensures
$
\regone = \scO\left(L|X|\sqrt{|A|T\ln\left(\frac{T|X||A|}{\delta}\right)}\right).
$
\end{lemma}
\begin{proof}
Since all losses are in $[0,1]$, we have
$
\regone \leq \sum_{t=1}^{T}\sum_{x,a} |\qhat_t(x,a) - q_t(x,a)|
= \sum_{t=1}^{T}\sum_{x,a} |q^{P_t^x,\pi_t}(x,a) - q_t(x,a)|,
$
where we define $P_t^x = P^{\qhat_t} \in \calP_{i_t}$ for all $x$ so that $\qhat_t = q^{P_t, \pi_t}$ (by the definition of $\pi_t$ and Lemma~\ref{lem:occupancy_measure}).
Applying Lemma~\ref{lem:key_lemma} finishes the proof.
\end{proof}

Note that in the proof above, we set $P_t^x$ to be the same for all $x$.
In fact, in this case our Lemma~\ref{lem:key_lemma} is similar to~\citep[Lemmas~B.2 and~B.3]{rosenberg19a} and it also suffices to use their looser confidence bound.
However, in the next application of Lemma~\ref{lem:key_lemma} to bounding $\regtwo$, it turns out to be critical to set $P_t^x$ to be different for different $x$ and also to use our tighter confidence bound.

\paragraph{Bounding $\regtwo$.}
To bound the term $\regtwo = \sum_{t=1}^T \langle \qhat_t , \ell_t - \ellhat_t \rangle$, we need to show that $\ellhat_t$ is not underestimating $\ell_t$ by too much, which, at a high-level, is also ensured due to the fact that the confidence set becomes more and more accurate for frequently visited state-action pairs.

\begin{lemma}
\label{lem:bias_1}
With probability at least $1-7\delta$, \alg ensures
\[
\regtwo = \scO\left(L|X|\sqrt{|A|T\ln\left(\frac{T|X||A|}{\delta}\right)}+\gamma |X||A|T\right).
\]
\end{lemma}

\begin{proof}
First note that $\langle \qhat_t, \ellhat_t \rangle$ is in $[0, L]$ because $P^{\qhat_t}\in \calP_{i_t}$ by the definition of $\qhat_t$ and thus $\qhat_t(x,a)\leq u_t(x, a)$ by the definition of $u_t$, which implies
\[
\sum_{x,a} \qhat_t(x,a)\ellhat_t(x,a) \leq \sum_{x,a}\ind\{x_{k(x)}=x, a_{k(x)}=a\} = L.
\]
Applying Azuma's inequality we thus have with probability at least $1-\delta$,
$
\sum_{t=1}^T \langle \qhat_t, \E_t[\ellhat_t] - \ellhat_t \rangle
\leq L\sqrt{2T\ln\frac{1}{\delta}}.
$
Therefore, we can bound $\regtwo$ by $\sum_{t=1}^T \langle \qhat_t , \ell_t - \E_t[\ellhat_t] \rangle + L\sqrt{2T\ln\tfrac{1}{\delta}}$ under this event.
We then focus on the term $\sum_t \langle \qhat_t , \ell_t - \E_t[\ellhat_t] \rangle$ and rewrite it as (by the definition of $\ellhat_t$)
\begin{align*}
&\sum_{t,x,a} \qhat_t(x,a)\ell_t(x,a)\left(1 - \frac{\E_t[\ind\{x_{k(x)}=x, a_{k(x)}=a\}]}{u_t(x,a)+\gamma}\right) \\
&= \sum_{t,x,a} \qhat_t(x,a)\ell_t(x,a)\left(1 - \frac{q_t(x,a)}{u_t(x,a)+\gamma}\right) \\
&=\sum_{t,x,a} \frac{\qhat_t(x,a)}{u_t(x,a)+\gamma}\left(u_t(x,a) - q_t(x,a) + \gamma\right) \\
&\leq \sum_{t,x,a} |u_t(x,a) - q_t(x,a)| + \gamma |X||A|T
\end{align*}
where the last step is again due to $\qhat_t(x,a) \leq u_t(x,a)$.
Finally, note that by Eq.~\eqref{eqn:UOB}, one has $u_t = q^{P_t^x,\pi_t}$ for $P_t^x = \argmax_{\Phat\in\calP_{i_t}} q^{\Phat,\pi_t}(x)$ (which is $\calF_t$-measurable and belongs to $\calP_{i_t}$ clearly).
Applying Lemma~\ref{lem:key_lemma} together with a union bound then finishes the proof.
\end{proof}

We point out again that this is the only part that requires using our new confidence set.
With the looser one used in previous work we can only show $\sum_{t,x,a} |u_t(x,a) - q_t(x,a)| = \scO\Big(L|X|^2\sqrt{|A|T\ln\big(\frac{T|X||A|}{\delta}\big)}\Big)$, with an extra $|X|$ factor.

% !TEX root = main.tex

\section{Conclusion}
In this work, we propose the first efficient algorithm with $\order(\sqrt{T})$ regret for learning MDPs with unknown transition function, adversarial losses, and bandit feedback.
Our main algorithmic contribution is to propose a tighter confidence bound together with a novel optimistic loss estimator based on upper occupancy bounds.
One natural open problem in this direction is to close the gap between our regret upper bound $\order(L|X|\sqrt{|A|T})$ and the lower bound of $\Omega(L\sqrt{|X||A|T})$~\citep{jin2018q}, which exists even for the full-information setting.

\section*{Acknowledgments}
HL is supported by NSF Awards IIS-1755781 and IIS-1943607. SS is partially supported by NSF-BIGDATA Award IIS-1741341 and an NSF-CAREER grant Award IIS-1846088. TY is partially supported by NSF BIGDATA grant IIS-1741341.

\bibliography{ref}

\begin{thebibliography}{34}
\providecommand{\natexlab}[1]{#1}
\providecommand{\url}[1]{\texttt{#1}}
\expandafter\ifx\csname urlstyle\endcsname\relax
  \providecommand{\doi}[1]{doi: #1}\else
  \providecommand{\doi}{doi: \begingroup \urlstyle{rm}\Url}\fi

\bibitem[Abbasi-Yadkori et~al.(2011)Abbasi-Yadkori, P{\'a}l, and
  Szepesv{\'a}ri]{abbasi2011improved}
Abbasi-Yadkori, Y., P{\'a}l, D., and Szepesv{\'a}ri, C.
\newblock Improved algorithms for linear stochastic bandits.
\newblock In \emph{Proceedings of the 24th International Conference on Neural
  Information Processing Systems}, pp.\  2312--2320, 2011.

\bibitem[Abernethy et~al.(2008)Abernethy, Hazan, and
  Rakhlin]{abernethy2008competing}
Abernethy, J.~D., Hazan, E., and Rakhlin, A.
\newblock Competing in the dark: An efficient algorithm for bandit linear
  optimization.
\newblock In \emph{Proceedings of the 21st Annual Conference on Learning
  Theory}, pp.\  263--274, 2008.

\bibitem[Allenberg et~al.(2006)Allenberg, Auer, Gy{\"o}rfi, and
  Ottucs{\'a}k]{AllenbergAuGyOt06}
Allenberg, C., Auer, P., Gy{\"o}rfi, L., and Ottucs{\'a}k, G.
\newblock Hannan consistency in on-line learning in case of unbounded losses
  under partial monitoring.
\newblock In \emph{Proceedings of the 17th international conference on
  Algorithmic Learning Theory}, pp.\  229--243, 2006.

\bibitem[Altman(1999)]{altman1999constrained}
Altman, E.
\newblock \emph{Constrained Markov decision processes}, volume~7.
\newblock CRC Press, 1999.

\bibitem[Arora et~al.(2012)Arora, Dekel, and Tewari]{arora2012deterministic}
Arora, R., Dekel, O., and Tewari, A.
\newblock Deterministic mdps with adversarial rewards and bandit feedback.
\newblock In \emph{Proceedings of the 28th Conference on Uncertainty in
  Artificial Intelligence}, pp.\  93--101, 2012.

\bibitem[Auer et~al.(2002{\natexlab{a}})Auer, Cesa-Bianchi, and
  Fischer]{auer2002finite}
Auer, P., Cesa-Bianchi, N., and Fischer, P.
\newblock Finite-time analysis of the multiarmed bandit problem.
\newblock \emph{Machine learning}, 47\penalty0 (2-3):\penalty0 235--256,
  2002{\natexlab{a}}.

\bibitem[Auer et~al.(2002{\natexlab{b}})Auer, Cesa-Bianchi, Freund, and
  Schapire]{auer2002nonstochastic}
Auer, P., Cesa-Bianchi, N., Freund, Y., and Schapire, R.~E.
\newblock The nonstochastic multiarmed bandit problem.
\newblock \emph{SIAM Journal on Computing}, 32\penalty0 (1),
  2002{\natexlab{b}}.

\bibitem[Azar et~al.(2017)Azar, Osband, and Munos]{azar2017minimax}
Azar, M.~G., Osband, I., and Munos, R.
\newblock Minimax regret bounds for reinforcement learning.
\newblock In \emph{Proceedings of the 34th International Conference on Machine
  Learning}, pp.\  263--272, 2017.

\bibitem[Beygelzimer et~al.(2011)Beygelzimer, Langford, Li, Reyzin, and
  Schapire]{beygelzimer2011contextual}
Beygelzimer, A., Langford, J., Li, L., Reyzin, L., and Schapire, R.
\newblock Contextual bandit algorithms with supervised learning guarantees.
\newblock In \emph{Proceedings of the Fourteenth International Conference on
  Artificial Intelligence and Statistics}, pp.\  19--26, 2011.

\bibitem[Burnetas \& Katehakis(1997)Burnetas and
  Katehakis]{burnetas1997optimal}
Burnetas, A.~N. and Katehakis, M.~N.
\newblock Optimal adaptive policies for markov decision processes.
\newblock \emph{Mathematics of Operations Research}, 22\penalty0 (1):\penalty0
  222--255, 1997.

\bibitem[Cheung et~al.(2019)Cheung, Simchi-Levi, and
  Zhu]{cheung2019reinforcement}
Cheung, W.~C., Simchi-Levi, D., and Zhu, R.
\newblock Reinforcement learning under drift.
\newblock \emph{arXiv preprint arXiv:1906.02922}, 2019.

\bibitem[Chu et~al.(2011)Chu, Li, Reyzin, and Schapire]{chu2011contextual}
Chu, W., Li, L., Reyzin, L., and Schapire, R.
\newblock Contextual bandits with linear payoff functions.
\newblock In \emph{Proceedings of the 14th International Conference on
  Artificial Intelligence and Statistics}, pp.\  208--214, 2011.

\bibitem[Dekel \& Hazan(2013)Dekel and Hazan]{dekel2013better}
Dekel, O. and Hazan, E.
\newblock Better rates for any adversarial deterministic mdp.
\newblock In \emph{Proceedings of the 30th International Conference on Machine
  Learning}, pp.\  675--683, 2013.

\bibitem[Dekel et~al.(2014)Dekel, Ding, Koren, and Peres]{dekel2014bandits}
Dekel, O., Ding, J., Koren, T., and Peres, Y.
\newblock Bandits with switching costs: ${T}^{2/3}$ regret.
\newblock In \emph{Proceedings of the 46th annual ACM symposium on Theory of
  computing}, pp.\  459--467, 2014.

\bibitem[Even-Dar et~al.(2009)Even-Dar, Kakade, and
  Mansour]{Even-Dar:2009:OMD:1599452.1599466}
Even-Dar, E., Kakade, S.~M., and Mansour, Y.
\newblock Online markov decision processes.
\newblock \emph{Mathematics of Operations Research}, 34\penalty0 (3):\penalty0
  726--736, 2009.

\bibitem[Fruit et~al.(2018)Fruit, Pirotta, Lazaric, and
  Ortner]{fruit2018efficient}
Fruit, R., Pirotta, M., Lazaric, A., and Ortner, R.
\newblock Efficient bias-span-constrained exploration-exploitation in
  reinforcement learning.
\newblock In \emph{Proceedings of the 35th International Conference on Machine
  Learning}, pp.\  1578--1586, 2018.

\bibitem[Hazan et~al.(2016)]{hazan2016introduction}
Hazan, E. et~al.
\newblock Introduction to online convex optimization.
\newblock \emph{Foundations and Trends{\textregistered} in Optimization},
  2\penalty0 (3-4):\penalty0 157--325, 2016.

\bibitem[Jaksch et~al.(2010)Jaksch, Ortner, and Auer]{jaksch2010near}
Jaksch, T., Ortner, R., and Auer, P.
\newblock Near-optimal regret bounds for reinforcement learning.
\newblock \emph{Journal of Machine Learning Research}, 11\penalty0
  (Apr):\penalty0 1563--1600, 2010.

\bibitem[Jin et~al.(2018)Jin, Allen-Zhu, Bubeck, and Jordan]{jin2018q}
Jin, C., Allen-Zhu, Z., Bubeck, S., and Jordan, M.~I.
\newblock Is q-learning provably efficient?
\newblock In \emph{Proceedings of the 32nd International Conference on Neural
  Information Processing Systems}, pp.\  4868--4878, 2018.

\bibitem[Lykouris et~al.(2019)Lykouris, Simchowitz, Slivkins, and
  Sun]{lykouris2019corruption}
Lykouris, T., Simchowitz, M., Slivkins, A., and Sun, W.
\newblock Corruption robust exploration in episodic reinforcement learning.
\newblock \emph{arXiv preprint arXiv:1911.08689}, 2019.

\bibitem[Maurer \& Pontil(2009)Maurer and Pontil]{maurer2009empirical}
Maurer, A. and Pontil, M.
\newblock Empirical bernstein bounds and sample variance penalization.
\newblock In \emph{Proceedings of the 22nd Annual Conference on Learning
  Theory}, 2009.

\bibitem[Neu(2015)]{neu2015explore}
Neu, G.
\newblock Explore no more: Improved high-probability regret bounds for
  non-stochastic bandits.
\newblock In \emph{Advances in Neural Information Processing Systems}, pp.\
  3168--3176, 2015.

\bibitem[Neu et~al.(2010)Neu, Gy{\"o}rgy, and Szepesv{\'a}ri]{neu2010online}
Neu, G., Gy{\"o}rgy, A., and Szepesv{\'a}ri, C.
\newblock The online loop-free stochastic shortest-path problem.
\newblock In \emph{Proceedings of the 23rd Annual Conference on Learning
  Theory}, pp.\  231--243, 2010.

\bibitem[Neu et~al.(2012)Neu, Gyorgy, and Szepesvari]{neu2012unknown}
Neu, G., Gyorgy, A., and Szepesvari, C.
\newblock The adversarial stochastic shortest path problem with unknown
  transition probabilities.
\newblock In \emph{Proceedings of the 15th International Conference on
  Artificial Intelligence and Statistics}, pp.\  805--813, 2012.

\bibitem[Neu et~al.(2014)Neu, Antos, Gy{\"o}rgy, and
  Szepesv{\'a}ri]{neu2014online}
Neu, G., Antos, A., Gy{\"o}rgy, A., and Szepesv{\'a}ri, C.
\newblock Online markov decision processes under bandit feedback.
\newblock \emph{IEEE Transactions on Automatic Control}, pp.\  676 -- 691,
  2014.

\bibitem[Ouyang et~al.(2017)Ouyang, Gagrani, Nayyar, and
  Jain]{ouyang2017learning}
Ouyang, Y., Gagrani, M., Nayyar, A., and Jain, R.
\newblock Learning unknown markov decision processes: a thompson sampling
  approach.
\newblock In \emph{Proceedings of the 31st International Conference on Neural
  Information Processing Systems}, pp.\  1333--1342, 2017.

\bibitem[Rosenberg \& Mansour(2019{\natexlab{a}})Rosenberg and
  Mansour]{rosenberg19a}
Rosenberg, A. and Mansour, Y.
\newblock Online convex optimization in adversarial {M}arkov decision
  processes.
\newblock In \emph{Proceedings of the 36th International Conference on Machine
  Learning}, pp.\  5478--5486, 2019{\natexlab{a}}.

\bibitem[Rosenberg \& Mansour(2019{\natexlab{b}})Rosenberg and
  Mansour]{rosenberg2019online}
Rosenberg, A. and Mansour, Y.
\newblock Online stochastic shortest path with bandit feedback and unknown
  transition function.
\newblock In \emph{Advances in Neural Information Processing Systems},
  2019{\natexlab{b}}.

\bibitem[Wang et~al.(2019)Wang, Dong, Chen, and Wang]{dong2019q}
Wang, Y., Dong, K., Chen, X., and Wang, L.
\newblock Q-learning with ucb exploration is sample efficient for
  infinite-horizon mdp.
\newblock In \emph{International Conference on Learning Representations}, 2019.

\bibitem[Wei et~al.(2019)Wei, Jafarnia-Jahromi, Luo, Sharma, and
  Jain]{wei2019model}
Wei, C.-Y., Jafarnia-Jahromi, M., Luo, H., Sharma, H., and Jain, R.
\newblock Model-free reinforcement learning in infinite-horizon average-reward
  markov decision processes.
\newblock \emph{arXiv preprint arXiv:1910.07072}, 2019.

\bibitem[Yu \& Mannor(2009)Yu and Mannor]{yu2009arbitrarily}
Yu, J.~Y. and Mannor, S.
\newblock Arbitrarily modulated markov decision processes.
\newblock In \emph{Proceedings of the 48h IEEE Conference on Decision and
  Control}, pp.\  2946--2953, 2009.

\bibitem[Yu et~al.(2009)Yu, Mannor, and Shimkin]{yu2009markov}
Yu, J.~Y., Mannor, S., and Shimkin, N.
\newblock Markov decision processes with arbitrary reward processes.
\newblock \emph{Mathematics of Operations Research}, 34\penalty0 (3):\penalty0
  737--757, 2009.

\bibitem[Zhang \& Ji(2019)Zhang and Ji]{zhang2019regret}
Zhang, Z. and Ji, X.
\newblock Regret minimization for reinforcement learning by evaluating the
  optimal bias function.
\newblock In \emph{Advances in Neural Information Processing Systems}, 2019.

\bibitem[Zimin \& Neu(2013)Zimin and Neu]{zimin2013}
Zimin, A. and Neu, G.
\newblock Online learning in episodic markovian decision processes by relative
  entropy policy search.
\newblock In \emph{Proceedings of the 26th International Conference on Neural
  Information Processing Systems}, pp.\  1583--1591, 2013.

\end{thebibliography}
\bibliographystyle{icml2020}

\newpage
% !TEX root = main.tex

\onecolumn
\appendix

\section{Omitted Details for the Algorithm}
In this section, we provide omitted details on how to implement our algorithm efficiently.

\subsection{Updating Occupancy Measure}
\label{app:proj}

This subsection explains how to implement the update defined in Eq.~\eqref{eqn:update_q} efficiently. We use almost the same approach as in \cite{rosenberg19a} with the only difference being the choice of confidence set. We provide details of the modification here for completeness. It has been shown in \cite{rosenberg19a} that Eq.~\eqref{eqn:update_q} can be decomposed into two steps: (1) compute $\tilde{q}_{t+1}(x, a, x')=\qhat_{t}(x, a, x')\exp\{-\eta\ellhat_t(x, a)\}$ for any $(x, a, x')$, which is the optimal solution of the unconstrained problem; (2) compute the projection step:
\begin{equation}\label{eqn:project_update}
\qhat_{t+1} = \argmin_{q\in\Delta(\calP_i)}\; D(q \;\|\; \tilde{q}_{t+1}),
\end{equation}
Since our choice of confidence set $\Delta(\calP_i)$ is different, the main change lies in the second step, whose constraint set can be written explicitly using the following set of linear equations:

\begin{align}
  % \label{normalize}
  &\forall k: &&\sum_{x\in X_k,a\in A,x'\in X_{k+1}}{q\left( x,a,x' \right)}=1, \nonumber
  \\
  &\forall k,\,\,\forall x\in X_k: &&\sum_{a\in A,x'\in X_{k+1}}{q\left( x,a,x' \right)}=\sum_{x'\in X_{k-1},a\in A}{q\left( x',a,x \right)},
  \nonumber
  \\
  &\forall k,\,\,\forall \left( x,a,x' \right) \in X_k\times A\times X_{k+1}: &&q\left( x,a,x' \right) \le \left[ \bar{P}_i\left( x'|x,a \right) +\epsilon _i\left( x'|x,a \right) \right] \sum_{y\in X_{k+1}}{q\left( x,a,y \right)},
  \nonumber
  \\
  & &&q\left( x,a,x' \right) \ge \left[ \bar{P}_i\left( x'|x,a \right) -\epsilon_i\left( x'|x,a \right) \right] \sum_{y\in X_{k+1}}{q\left( x,a,y \right)},
  \nonumber
  \\
  \label{non-negative}
  & &&q\left( x,a,x' \right) \ge 0.
\end{align}

Therefore, the projection step Eq.~\eqref{eqn:project_update} is a convex optimization problem with linear constraints, which can be solved in polynomial time. This optimization problem can be further reformulated into a dual problem, which is a convex optimization problem with only non-negativity constraints, and thus can be solved more efficiently.

\begin{lemma}
The dual problem of Eq.\eqref{eqn:project_update} is to solve
\begin{equation*}
\mu _t,\beta _t= \argmin_{\mu ,\beta \ge 0} \sum_{k=0}^{L-1}{\ln Z_{t}^{k}\left( \mu ,\beta \right)}
\end{equation*}
where $\beta := \{\beta(x)\}_x$ and $\mu := \{\mu^+(x, a, x'), \mu^-(x, a, x')\}_{(x, a, x')}$ are dual variables and 
\begin{align*}
   Z_{t}^{k}\left( \mu ,\beta \right) &= \sum_{x\in X_k,a\in A,x'\in X_{k+1}}{\qhat_t\left( x,a,x' \right) \exp \left\{ B_{t}^{\mu ,\beta}\left( x,a,x' \right) \right\}},\\
  B_{t}^{\mu ,\beta}\left( x,a,x' \right) &= \beta \left( x' \right) -\beta \left( x \right) +\left( \mu ^--\mu ^+ \right) \left( x,a,x' \right) -\eta \ellhat_t\left( x,a \right)
  \\
  +&\sum_{y\in X_{k\left( x \right) +1}}{\left( \mu ^+-\mu ^- \right) \left( x,a,y \right) \bar{P}_i\left( y|x,a \right) +\left( \mu ^++\mu ^- \right) \left( x,a,y \right) \epsilon _i\left( y|x,a \right)}.
\end{align*}
Furthermore, the optimal solution to Eq.\eqref{eqn:project_update} is given by
\begin{equation*}
   % \label{projection-form}
\qhat_{t+1}\left( x,a,x' \right) =\frac{\qhat_t\left( x,a,x' \right)}{Z_{t}^{k\left( x \right)}\left( \mu _t,\beta _t \right)}\exp \left\{ B_{t}^{\mu _t,\beta _t}\left( x,a,x' \right) \right\}.
 \end{equation*}
\end{lemma}

\begin{proof}
% The primal problem can also be solved using a dual approach as developed in Appendix A of \cite{rosenberg19a}, of which we will sketch the derivation below. We write the Lagrangian.

In the following proof, we omit the non-negativity constraint Eq.~\eqref{non-negative}. This is without loss of generality, since the optimal solution for the modified version of Eq.\eqref{eqn:project_update} without the non-negativity constraint Eq.~\eqref{non-negative} turns out to always satisfy the non-negativity constraint.

 % because it will always be satisfied following our algorithm by \eqref{projection-form} and the non-negativity of $\qhat_0$. So we are actually solving \eqref{eqn:project_update} using a larger feasible set but the optimal solution remains the same.

% Therefore, the projection step Eq.\eqref{eqn:project_update} is a convex optimization problem with linear constraints. It is known that such a problem can be solved efficiently. The primal problem can also be solved using a dual approach as developed in Appendix A of \cite{rosenberg19a}, of which we will sketch the derivation below. 
We write the Lagrangian as:
\begin{align*}
  \mathcal{L}\left( q, \lambda, \beta, \mu \right) =&D\left( q||\tilde{q}_{t+1} \right) +\sum_{k=0}^{L-1}{\lambda _k\left( \sum_{x\in X_k,a\in A,x'\in X_{k+1}}{q\left( x,a,x' \right)}-1 \right)}
  \\
  &+\sum_{k=1}^{L-1}{\sum_{x\in X_k}{\beta \left( x \right) \left( \sum_{a\in A,x'\in X_{k+1}}{q\left( x,a,x' \right)}-\sum_{x'\in X_{k-1},a\in A}{q\left( x',a,x \right)} \right)}}
  \\
  &+\sum_{k=0}^{L-1}{\sum_{x\in X_k,a\in A,x'\in X_{k+1}}{\mu ^+\left( x,a,x' \right) \left( q\left( x,a,x' \right) -\left[ \bar{P}_i\left( x'|x,a \right) +\epsilon _i\left( x'|x,a \right) \right] \sum_{y\in X_{k+1}}{q\left( x,a,y \right)} \right)}}
  \\
  &+\sum_{k=0}^{L-1}{\sum_{x\in X_k,a\in A,x'\in X_{k+1}}{\mu ^-\left( x,a,x' \right) \left( \left[ \bar{P}_i\left( x'|x,a \right) -\epsilon _i\left( x'|x,a \right) \right] \sum_{y\in X_{k+1}}{q\left( x,a,y \right)}-q\left( x,a,x' \right) \right)}}
\end{align*}
%<<<<<<< HEAD
%where $\lambda ,\beta ,\mu =\left( \mu ^+,\mu ^- \right)$ are Lagrange multipliers. We also define $\beta \left( x_0 \right) =\beta \left( x_L \right) =0$ for convenience. Now taking the derivative we have
%=======
where $\lambda:= \{\lambda_k\}_k$, $\beta := \{\beta(x)\}_x$ and $\mu := \{\mu^+(x, a, x'), \mu^-(x, a, x')\}_{(x, a, x')}$ are Lagrange multipliers. We also define $\beta \left( x_0 \right) =\beta \left( x_L \right) =0$ for convenience. 
Now taking the derivative we have
\begin{align*}
  \frac{\partial \mathcal{L}}{\partial q\left( x,a,x' \right)}=&\ln q\left( x,a,x' \right) -\ln \tilde{q}_{t+1}\left( x,a,x' \right) +\lambda _{k\left( x \right)}+\beta \left( x \right) -\beta \left( x' \right) +\left( \mu ^+-\mu ^- \right) \left( x,a,x' \right)
  \\
  &-\sum_{y\in X_{k\left( x \right) +1}}{\left( \mu ^+-\mu ^- \right) \left( x,a,y \right) \bar{P}_i\left( y|x,a \right) +\left( \mu ^++\mu ^- \right) \left( x,a,y \right) \epsilon _i\left( y|x,a \right)}
  \\
  =&\ln q\left( x,a,x' \right) -\ln \tilde{q}_{t+1}\left( x,a,x' \right) +\lambda _{k\left( x \right)}-\eta \ellhat_t\left( x,a \right) -B_{t}^{\mu ,\beta}\left( x,a,x' \right).
\end{align*}

Setting the derivative to zero gives the explicit form of the optimal $q^\star$ by
\begin{align*}
  q^\star\left( x,a,x' \right) &=\tilde{q}_{t+1}\left( x,a,x' \right) \exp \left\{ -\lambda _{k\left( x \right)}+\eta \ellhat_t\left( x,a \right) +B_{t}^{\mu ,\beta}\left( x,a,x' \right) \right\}
  \\
  &=\qhat_t\left( x,a,x' \right) \exp \left\{ -\lambda _{k\left( x \right)}+B_{t}^{\mu ,\beta}\left( x,a,x' \right) \right\}.
\end{align*}

On the other hand, setting $\partial \mathcal{L}/\partial \lambda_k = 0$ shows that the optimal $\lambda^\star$ satisfies
$$
\exp \left\{ \lambda^\star_k \right\} =\sum_{x\in X_k,a\in A,x'\in X_{k+1}}{\qhat_t\left( x,a,x' \right) \exp \left\{ B_{t}^{\mu ,\beta}\left( x,a,x' \right) \right\}} = Z_{t}^{k}\left( \mu ,\beta \right).
$$
% Thus we can define the normalizer
%  \begin{align*}
%    Z_{t}^{k}\left( \mu ,\beta \right) =\sum_{x\in X_k,a\in A,x'\in X_{k+1}}{\hat{q}_t\left( x,a,x' \right) \exp \left\{ B_{t}^{\mu ,\beta}\left( x,a,x' \right) \right\}},
%  \end{align*}
It is straightforward to check that strong duality holds, and thus the optimal dual variables $\mu^\star, \beta^\star$ are given by
\begin{equation*}
  \mu^\star,\beta^\star= \argmax_{\mu ,\beta \ge 0} \max_{\lambda} \min_q \mathcal{L}\left( q, \lambda, \beta, \mu \right)
  =\argmax_{\mu ,\beta \ge 0} \mathcal{L}\left( q^\star, \lambda^\star, \beta, \mu \right).
\end{equation*}  
Finally, we note the equality
\begin{align*}
\mathcal{L}\left( q, \lambda, \beta, \mu \right) =& D\left( q||\tilde{q}_{t+1} \right)  + \sum_{k=0}^{L-1} \sum_{x\in X_k,a\in A,x'\in X_{k+1}} \left(\frac{\partial \mathcal{L}}{\partial q\left( x,a,x' \right)} - \ln q\left( x,a,x' \right) +\ln \tilde{q}_{t+1}\left( x,a,x' \right)\right) q(x, a, x') - \sum_{k=1}^{L-1} \lambda_k\\
=& \sum_{k=0}^{L-1} \sum_{x\in X_k,a\in A,x'\in X_{k+1}} \left[\left(\frac{\partial \mathcal{L}}{\partial q\left( x,a,x' \right) } - 1  \right) q(x, a, x') + \tilde{q}_{t+1}(x, a, x')\right] - \sum_{k=1}^{L-1} \lambda_k.
\end{align*}

This, combined with the fact that $q^\star$ has zero partial derivative, gives
\begin{align*}
\mathcal{L}\left( q^\star, \lambda^\star, \beta, \mu \right)
% = & \sum_{k=0}^{L-1} \sum_{x\in X_k,a\in A,x'\in X_{k+1}} \left[ -q^\star(x, a, x') + \tilde{q}_{t+1}(x, a, x')\right] - \sum_{k=1}^{L-1} \lambda^\star_k\\
= & -L + \sum_{k=0}^{L-1} \sum_{x\in X_k,a\in A,x'\in X_{k+1}}\tilde{q}_{t+1}(x, a, x') - \sum_{k=0}^{L-1}{\ln Z_{t}^{k}\left( \mu ,\beta \right)}.
\end{align*}
Note that the first two terms in the last expression are independent of $(\mu, \beta)$. We thus have:
\begin{equation*}
\mu^\star,\beta^\star  =\argmax_{\mu ,\beta \ge 0} \mathcal{L}\left( q^\star, \lambda^\star, \beta, \mu \right)
= \argmin_{\mu ,\beta \ge 0} \sum_{k=0}^{L-1}{\ln Z_{t}^{k}\left( \mu ,\beta \right)}.
\end{equation*}
Combining all equations for $(q^\star, \lambda^\star, \mu^\star, \beta^\star)$ finishes the proof.
% =\argmin_{\mu ,\beta \ge 0}\sum_{k=0}^{L-1}{\ln Z_{t}^{k}\left( \mu ,\beta \right)}.
% \end{align*}
\end{proof}

% Therefore, the solution must be of the following form
%  \begin{equation}
%    % \label{projection-form}
% \hat{q}_{t+1}\left( x,a,x' \right) =\frac{\hat{q}_t\left( x,a,x' \right)}{Z_{t}^{k\left( x \right)}\left( \mu _t,\beta _t \right)}\exp \left\{ B_{t}^{\mu _t,\beta _t}\left( x,a,x' \right) \right\}  ,
%  \end{equation}
%  where $\mu_t,\beta_t $ are dual variables given by solving the dual problem
% \begin{align*}
%   \mu _t,\beta _t=&\argmax_{\mu ,\beta \ge 0} \max_{\lambda} \min_q \mathcal{L}\left( q, \lambda, \beta, \mu \right)
%   \\
%   =&\argmax_{\mu ,\beta \ge 0} \mathcal{L}\left( \hat{q}_{t+1}, \hat{\lambda}, \beta, \mu \right)
%   \\
%   =&\argmin_{\mu ,\beta \ge 0}\sum_{k=0}^{L-1}{\ln Z_{t}^{k}\left( \mu ,\beta \right)}
% \end{align*}
% Finally we just need to justify the strong duality condition. This is because
% \begin{enumerate}
%   \item The target function of the primal problem is always convex and bounded from below by definition of KL-divergence.
%   \item All the constrains are linear.
%   \item Slater's condition hold. For example, check the occupancy measure generated by any policy $\pi$ and the transition kernel $\bar{P}_i\left( x'|x,a \right) $.
% \end{enumerate}

% The advantage of the dual approach is that the constrains are significantly simplified (and thus projection onto feasible set is straightforward) and it provides some information about the form of the solution.

\subsection{Computing Upper Occupancy Bounds}
\label{app:alg}
This subsection explains how to greedily solve the following optimization problem from Eq.~\eqref{eqn:DP}:
\[
\max_{\Phat(\cdot|\xtilde,a)} \sum_{x' \in X_{k(\xtilde)+1}} \Phat(x' | \xtilde, a) f(x')
\]
subject to $\Phat(\cdot|\xtilde,a)$ being a valid distribution over $X_{k(\xtilde)+1}$ and for all $x' \in X_{k(\xtilde)+1}$,
\[
\left| \Phat(x'|\xtilde,a) - \bar{P}_i(x'|\xtilde,a) \right| \leq \epsilon_i(x'|\xtilde,a),
\]
where $(\xtilde,a)$ is some fixed state-action pair, $\epsilon_i(x'|\xtilde,a)$ is defined in Eq.~\eqref{eqn:width}, and the value of $f(x')$ for any $x' \in X_{k(\xtilde)+1}$ is known.
To simplify notation, let $n = |X_{k(\xtilde)+1}|$, and $\sigma: [n] \rightarrow X_{k(\xtilde)+1}$ be a bijection such that
\[
f(\sigma(1)) \leq f(\sigma(2)) \leq \cdots \leq f(\sigma(n)).
\]
Further let $\bar{p}$ and $\epsilon$ be shorthands of $\bar{P}_i(\cdot|\xtilde,a)$ and $\epsilon_i(\cdot|\xtilde,a)$ respectively.
With these notations, the problem becomes
\[
\max_{\substack{p \in \fR_+^n: \sum_{x'} p(x') = 1 \\ |p(x') -  \bar{p}(x')| \leq \epsilon(x')}} \;\;\sum_{j=1}^n p(\sigma(j))f(\sigma(j)).
\]
Clearly, the maximum is achieved by redistributing the distribution $\bar{p}$ so that it puts as much weight as possible on states with large $f$ value under the constraint.
%
%This can be implemented efficiently by starting with $p = \bar{p}$,
%then making two passes of the $n$ states:
%1) in the first pass, iterates $x$ from $\sigma(n)$ to $\sigma(1)$, increase $p(x)$ as much as possible until it reaches $1$ or until a total amount of $\epsilon/2$ has been increased;
%2) in the second pass, iterates $x$ from $\sigma(1)$ to $\sigma(n)$, decrease $p(x)$ as much as possible until it reaches $0$ or until $p$ becomes a distribution again.
%
This can be implemented efficiently by maintaining two pointers $j^-$ and $j^+$ starting from $1$ and $n$ respectively, and considering moving as much weight as possible from state $x^- = \sigma(j^-)$ to state $x^+ = \sigma(j^+)$.
More specifically, the maximum possible weight change for $x^-$ and $x^+$ are $\delta^- = \min\{\bar{p}(x^-),  \epsilon(x^-)\}$ and $\delta^+ = \min\{1 - \bar{p}(x^+),  \epsilon(x^+)\}$ respectively,
and thus we move $\min\{\delta^-, \delta^+\}$ amount of weight from $x^-$ to $x^+$.
In the case where $\delta^- \leq \delta^+$, no more weight can be decreased from $x^-$ and we increase the pointer $j^-$ by $1$ as well as decreasing $\epsilon(x^+)$ by $\delta^-$ to reflect the change in maximum possible weight increase for $x^+$.
The situation for the case $\delta^- > \delta^+$ is similar.
The procedure stops when the two pointers coincide.
See Algorithm~\ref{alg:greedy} for the complete pseudocode.

We point out that the step of sorting the values of $f$ and finding $\sigma$ can in fact be done only once for each layer (instead of every call of Algorithm~\ref{alg:greedy}).
For simplicity, we omit this refinement.

%\HLcomment{I made major changes to the algorithm and the description to make things easier to interpret. I don't think we need a proof for correctness.} \CJcomment{Agree, I think this is already pretty straight-forward.}

\begin{algorithm}[!htbp]
\caption{\textsc{Greedy}}
\label{alg:greedy}
\begin{algorithmic}
\STATE {\bfseries Input:} $f: X\rightarrow[0,1]$, a distribution $\bar{p}$ over $n$ states of layer $k$ , positive numbers $\{\epsilon(x)\}_{x \in X_k}$

\STATE {\bfseries Initialize:} $j^- = 1, j^+ = n$, sort $\{f(x)\}_{x\in X_k}$ and find $\sigma$ such that
$
f(\sigma(1)) \leq f(\sigma(2)) \leq \cdots \leq f(\sigma(n))
$

\item[]
\WHILE{$j^- < j^+$}
\STATE $x^- =\sigma(j^-), x^+ =\sigma(j^+)$
\STATE $\delta^{-} = \min\{\bar{p}(x^-),  \epsilon(x^-)\}$ \hfill$\rhd$maximum weight to decrease for state $x^-$
\STATE $\delta^{+} = \min\{1 - \bar{p}(x^+),  \epsilon(x^+)\}$ \hfill$\rhd$maximum weight to increase for state $x^+$
\STATE $\bar{p}(x^-) \leftarrow \bar{p}(x^-) - \min\{\delta^{-}, \delta^{+}\}$
\STATE $\bar{p}(x^+) \leftarrow \bar{p}(x^+) + \min\{\delta^{-}, \delta^{+}\}$
\IF{$\delta_{-} \leq \delta_{+}$}
\STATE $\epsilon(x^+) \leftarrow \epsilon(x^+) - \delta^{-}$
\STATE $j^- \leftarrow j^- + 1$
\ELSE
\STATE $\epsilon(x^-) \leftarrow \epsilon(x^-) - \delta^{+}$
\STATE $j^+ \leftarrow j^+ - 1$
\ENDIF
\ENDWHILE

\STATE {\bfseries Return:} $\sum_{j=1}^n \bar{p}(\sigma(j))f(\sigma(j))$
\end{algorithmic}
\end{algorithm}

\section{Omitted Details for the Analysis}
\label{app:analysis}

In this section, we provide omitted proofs for the regret analysis of our algorithm.

\subsection{Auxiliary Lemmas}\label{app:concentration}
First, we prove Lemma~\ref{lem:confidence_sets} which states that with probability at least $1-4\delta$, the true transition function $P$ is within the confidence set $\calP_i$ for all epoch $i$.

\begin{proof}[Proof of Lemma~\ref{lem:confidence_sets}]
By the empirical Bernstein inequality~\citep[Theorem 4]{maurer2009empirical} and union bounds, we have with probability at least $1-4\delta$, for all $(x,a,x')\in X_k \times A \times X_{k+1}$, $k=0, \ldots, L-1$, and any $i \le T$,
\begin{align*}
\left|P(x'|x,a) - \bar{P}_{i}(x'|x,a)\right|
&\leq \sqrt{\frac{2\bar{P}_{i}(x'|x,a)(1-\bar{P}_{i}(x'|x,a))\ln\left(\frac{T|X|^2|A|}{\delta}\right)}{\max\{1,N_i(x,a)-1\}}} + \frac{7\ln\left(\frac{T|X|^2|A|}{\delta}\right)}{3\max\{1,N_i(x,a)-1\}} \\
&\leq 2\sqrt{\frac{\bar{P}_{i}(x'|x,a)\ln\left(\frac{T|X||A|}{\delta}\right)}{\max\{1,N_i(x,a)-1\}}} + \frac{14\ln\left(\frac{T|X||A|}{\delta}\right)}{3\max\{1,N_i(x,a)-1\}} = \epsilon_i(x'|x,a)
\end{align*}
which finishes the proof.
\end{proof}

Next, we state three lemmas that are useful for the rest of the proof.
The first one shows a convenient bound on the difference between the true transition function and any transition function from the confidence set.

\begin{lemma}\label{lem:confidence_width}
Under the event of Lemma~\ref{lem:confidence_sets}, for all epoch $i$, all $\Phat \in \calP_i$, all $k = 0, \ldots, L-1$ and $(x,a,x')\in X_k \times A \times X_{k+1}$, we have
\[
\left|\Phat(x'|x,a) - P(x'|x,a)\right| = \scO\left(
\sqrt{\frac{P(x'|x,a)\ln\left(\frac{T|X||A|}{\delta}\right)}{\max\{1,N_i(x,a)\}}} + \frac{\ln\left(\frac{T|X||A|}{\delta}\right)}{\max\{1,N_i(x,a)\}}
\right) \triangleq \epsilon_i^\star(x'|x,a).
\]
\end{lemma}

\begin{proof}
Under the event of Lemma~\ref{lem:confidence_sets}, we have
\[
\bar{P}_{i}(x'|x,a) \leq P(x'|x,a) + 2\sqrt{\frac{\bar{P}_{i}(x'|x,a)\ln\left(\frac{T|X||A|}{\delta}\right)}{\max\{1,N_i(x,a)-1\}}} + \frac{14\ln\left(\frac{T|X||A|}{\delta}\right)}{3\max\{1,N_i(x,a)-1\}}.
\]
Viewing this as a quadratic inequality of $\sqrt{\bar{P}_{i}(x'|x,a)}$ and solving for $\bar{P}_{i}(x'|x,a)$ prove the lemma.
\end{proof}

The next one is a standard Bernstein-type concentration inequality for martingale.
We use the version from~\citep[Theorem~1]{beygelzimer2011contextual}.
\begin{lemma}
\label{lem:Freedman}
Let $Y_1, \ldots, Y_T$ be a martingale difference sequence with respect to a filtration $\calF_1, \ldots, \calF_T$. Assume $Y_t \leq R$ a.s. for all $i$. Then for any $\delta \in (0,1)$ and $\lambda \in [0, 1/R]$, with probability at least $1 - \delta$, we have
\[
\sum_{t=1}^T Y_t \leq \lambda \sum_{t=1}^T \E_t[Y_t^2] + \frac{\ln(1/\delta)}{\lambda}.
\]
\end{lemma}

The last one is a based on similar ideas used for proving many other optimistic algorithms.

\begin{lemma}\label{lem:aux}
With probability at least $1 - 2\delta$, we have for all $k = 0, \ldots, L-1$,
\begin{equation}\label{eqn:aux1}
\sum_{t=1}^T\sum_{x\in X_k, a\in A} \frac{q_t(x,a)}{\max\{1, N_{i_t}(x,a)\}} =
 \scO\left( |X_k||A|\ln T + \ln(L/\delta) \right)
\end{equation}
and
\begin{equation}\label{eqn:aux2}
\sum_{t=1}^T\sum_{x\in X_k, a\in A} \frac{q_t(x,a)}{\sqrt{\max\{1, N_{i_t}(x,a)\}}} =  \scO\left(\sqrt{|X_k||A| T}+ |X_k||A|\ln T + \ln(L/\delta) \right).
\end{equation}
\end{lemma}
\begin{proof}
Let $\ind_t(x,a)$ be the indicator of whether the pair $(x,a)$ is visited in episode $t$ so that $\E_t[\ind_t(x,a)] = q_t(x,a)$.
We decompose the first quantity as
\[
\sum_{t=1}^T\sum_{x\in X_k, a\in A} \frac{q_t(x,a)}{\max\{1, N_{i_t}(x,a)\}} =
\sum_{t=1}^T\sum_{x\in X_k, a\in A} \frac{\ind_t(x,a)}{\max\{1, N_{i_t}(x,a)\}}
+ \sum_{t=1}^T\sum_{x\in X_k, a\in A} \frac{q_t(x,a) - \ind_t(x,a)}{\max\{1, N_{i_t}(x,a)\}}.
\]
The first term can be bounded as
\begin{align*}
\sum_{x\in X_k, a\in A}\sum_{t=1}^T \frac{\ind_t(x,a)}{\max\{1, N_{i_t}(x,a)\}}
= \sum_{x\in X_k, a\in A} \scO\left( \ln T \right) = \scO\left( |X_k||A|\ln T \right).
\end{align*}
To bound the second term, we apply Lemma~\ref{lem:Freedman} with $Y_t = \sum_{x\in X_k, a\in A} \frac{q_t(x,a) - \ind_t(x,a)}{\max\{1, N_{i_t}(x,a)\}} \leq 1$, $\lambda = 1/2$, and the fact
\begin{align*}
\E_t[Y_t^2] &\leq \E_t\left[\left(\sum_{x\in X_k, a\in A} \frac{\ind_t(x,a)}{\max\{1, N_{i_t}(x,a)\}}\right)^2\right] \\
&= \E_t\left[\sum_{x\in X_k, a\in A} \frac{\ind_t(x,a)}{\max\{1, N_{i_t}^2(x,a)\}}\right] \tag{$\ind_t(x,a)\ind_t(x',a')=0$ for $x \neq x'\in X_k$} \\
&\leq \sum_{x\in X_k, a\in A} \frac{q_t(x,a)}{\max\{1, N_{i_t}(x,a)\}},
\end{align*}
which gives with probability at least $1 - \delta/L$,
\[
\sum_{t=1}^T\sum_{x\in X_k, a\in A} \frac{q_t(x,a) - \ind_t(x,a)}{\max\{1, N_{i_t}(x,a)\}}
\leq \frac{1}{2}\sum_{t=1}^T\sum_{x\in X_k, a\in A} \frac{q_t(x,a)}{\max\{1, N_{i_t}(x,a)\}} + 2\ln\left(\frac{L}{\delta}\right).
\]
Combining these two bounds, rearranging, and applying a union bound over $k$ prove Eq.~\eqref{eqn:aux1}.

Similarly, we decompose the second quantity as
\[
\sum_{t=1}^T\sum_{x\in X_k, a\in A} \frac{q_t(x,a)}{\sqrt{\max\{1, N_{i_t}(x,a)\}}} =
\sum_{t=1}^T\sum_{x\in X_k, a\in A} \frac{\ind_t(x,a)}{\sqrt{\max\{1, N_{i_t}(x,a)\}}}
+ \sum_{t=1}^T\sum_{x\in X_k, a\in A} \frac{q_t(x,a) - \ind_t(x,a)}{\sqrt{\max\{1, N_{i_t}(x,a)\}}}.
\]
The first term is bounded by
\begin{align*}
\sum_{x\in X_k, a\in A}\sum_{t=1}^T \frac{\ind_t(x,a)}{\sqrt{\max\{1, N_{i_t}(x,a)\}}}  &= \scO\left(\sum_{x\in X_k, a\in A}\sqrt{N_{i_T}(x,a)} \right) \\
&\leq \scO\left(\sqrt{|X_k||A|\sum_{x\in X_k, a\in A}N_{i_T}(x,a)} \right)
= \scO\left(\sqrt{|X_k||A|T} \right),
\end{align*}
where the second line uses the Cauchy-Schwarz inequality and the fact $\sum_{x\in X_k, a\in A}N_{i_T}(x,a) \leq T$.
To bound the second term, we again apply Lemma~\ref{lem:Freedman} with $Y_t = \sum_{x\in X_k, a\in A} \frac{q_t(x,a) - \ind_t(x,a)}{\sqrt{\max\{1, N_{i_t}(x,a)\}}} \leq 1$, $\lambda = 1$, and the fact
\begin{align*}
\E_t[Y_t^2] &\leq \E_t\left[\left(\sum_{x\in X_k, a\in A} \frac{\ind_t(x,a)}{\sqrt{\max\{1, N_{i_t}(x,a)\}}}\right)^2\right] =
\sum_{x\in X_k, a\in A} \frac{q_t(x,a)}{\max\{1, N_{i_t}(x,a)\}},
\end{align*}
which shows with probability at least $1 - \delta/L$,
\[
\sum_{t=1}^T\sum_{x\in X_k, a\in A} \frac{q_t(x,a) - \ind_t(x,a)}{\sqrt{\max\{1, N_{i_t}(x,a)\}}} \leq
\sum_{t=1}^T \sum_{x\in X_k, a\in A} \frac{q_t(x,a)}{\max\{1, N_{i_t}(x,a)\}} + \ln\left(\frac{L}{\delta}\right).
\]
Combining Eq.~\eqref{eqn:aux1} and a union bound proves Eq.~\eqref{eqn:aux2}.
\end{proof}

\subsection{Proof of the Key Lemma}
\label{app:key_lemma}
We are now ready to prove Lemma~\ref{lem:key_lemma}, the key lemma of our analysis which requires using our new confidence set.

\begin{proof}[Proof of Lemma~\ref{lem:key_lemma}]
To simplify notation, let $q_t^x = q^{P_t^x,\pi_t}$.
Note that for any occupancy measure $q$, by definition we have for any $(x,a)$ pair,
\[
q(x,a) = \pi^q(x|a)\sum_{\{x_k \in X_k, a_k \in A\}_{k=0}^{k(x)-1}}
% \sum_{k=0}^{k(x)-1}\sum_{x_k\in X_k, a_k \in A} 
\prod_{h=0}^{k(x)-1}\pi^q(a_h|x_h) \prod_{h=0}^{k(x)-1} P^q(x_{h+1} | x_h, a_h).
\]
where we define $x_{k(x)} = x$ for convenience.
Therefore, we have
\[
|q_t^x(x,a) - q_t(x,a)| = \pi_t(x|a)
% \sum_{k=0}^{k(x)-1}\sum_{x_k\in X_k, a_k \in A} 
\sum_{\{x_k, a_k\}_{k=0}^{k(x)-1}}
\prod_{h=0}^{k(x)-1}\pi_t(a_h|x_h)
\left(\prod_{h=0}^{k(x)-1} P_t^x(x_{h+1} | x_h, a_h)
- \prod_{h=0}^{k(x)-1} P(x_{h+1} | x_h, a_h)
\right).
\]
By adding and subtracting $k(x)-1$ terms we rewrite the last term in the parentheses as
\begin{align*}
&\prod_{h=0}^{k(x)-1} P_t^x(x_{h+1} | x_h, a_h)
- \prod_{h=0}^{k(x)-1} P(x_{h+1} | x_h, a_h) \\
&= \prod_{h=0}^{k(x)-1} P_t^x(x_{h+1} | x_h, a_h)
- \prod_{h=0}^{k(x)-1} P(x_{h+1} | x_h, a_h) \pm
\sum_{m=1}^{k(x)-1} \prod_{h=0}^{m-1} P(x_{h+1} | x_h, a_h) \prod_{h=m}^{k(x)-1} P_t^x(x_{h+1} | x_h, a_h) \\
&= \sum_{m=0}^{k(x)-1}\left(P_t^x(x_{m+1} | x_m, a_m) - P(x_{m+1} | x_m, a_m)\right) \prod_{h=0}^{m-1} P(x_{h+1} | x_h, a_h) \prod_{h=m+1}^{k(x)-1} P_t^x(x_{h+1} | x_h, a_h),
\end{align*}
which, by Lemma~\ref{lem:confidence_width}, is bounded by
\[
\sum_{m=0}^{k(x)-1}\epsilon_{i_t}^\star(x_{m+1}|x_m, a_m) \prod_{h=0}^{m-1} P(x_{h+1} | x_h, a_h) \prod_{h=m+1}^{k(x)-1} P_t^x(x_{h+1} | x_h, a_h).
\]
We have thus shown
% \CJcomment{Should we use $\sum_{x_1, a_1, \ldots, x_{k(x)-1}, a_{k(x)-1}}$ or $\sum_{\{x_k, a_k\}_{k=1}^{k(x)-1}}$ instead of $\sum_{k=0}^{k(x)-1}\sum_{x_k, a_k}$? In standard math, the latter does not mean the former two, and it is a bit confusing here.}
\begin{align}
&|q_t^x(x,a) - q_t(x,a)| \notag \\
&\leq \pi_t(x|a)
% \sum_{k=0}^{k(x)-1}\sum_{x_k, a_k} 
\sum_{\{x_k, a_k\}_{k=0}^{k(x)-1}}
\prod_{h=0}^{k(x)-1}\pi_t(a_h|x_h)
\sum_{m=0}^{k(x)-1}\epsilon_{i_t}^\star(x_{m+1}|x_m, a_m) \prod_{h=0}^{m-1} P(x_{h+1} | x_h, a_h) \prod_{h=m+1}^{k(x)-1} P_t^x(x_{h+1} | x_h, a_h) \notag\\
&= \sum_{m=0}^{k(x)-1}
% \sum_{k=0}^{k(x)-1}\sum_{x_k, a_k}
\sum_{\{x_k, a_k\}_{k=0}^{k(x)-1}}
\epsilon_{i_t}^\star(x_{m+1}|x_m, a_m) \left(\pi_t(a_m|x_m)\prod_{h=0}^{m-1} \pi_t(a_h|x_h)P(x_{h+1} | x_h, a_h)\right) \notag\\
&\hspace{20em}  \cdot \left(\pi_t(x|a)\prod_{h=m+1}^{k(x)-1} \pi_t(a_h|x_h) P_t^x(x_{h+1} | x_h, a_h) \right) \notag\\
&= \sum_{m=0}^{k(x)-1}\sum_{x_m, a_m, x_{m+1}}
\epsilon_{i_t}^\star(x_{m+1}|x_m, a_m) \left(
% \sum_{k=0}^{m-1}\sum_{x_k, a_k} 
\sum_{\{x_k, a_k\}_{k=0}^{m-1}}
\pi_t(a_m|x_m)\prod_{h=0}^{m-1} \pi_t(a_h|x_h)P(x_{h+1} | x_h, a_h)\right) \notag\\
&\hspace{15em}  \cdot \left(\sum_{a_{m+1}}
% \sum_{k=m+2}^{k(x)-1}\sum_{x_k, a_k} 
\sum_{\{x_k, a_k\}_{k=m+2}^{k(x)-1}}
\pi_t(x|a)\prod_{h=m+1}^{k(x)-1} \pi_t(a_h|x_h) P_t^x(x_{h+1} | x_h, a_h) \right) \notag\\
&= \sum_{m=0}^{k(x)-1}\sum_{x_m, a_m, x_{m+1}}
\epsilon_{i_t}^\star(x_{m+1}|x_m, a_m) q_t(x_m, a_m) q_t^x(x, a | x_{m+1}), \label{eqn:bound1}
\end{align}
where we use $q_t^x(x, a | x_{m+1})$ to denote the probability of encountering pair $(x,a)$ given that $x_{m+1}$ was visited in layer $m+1$, under policy $\pi_t$ and transition $P_t^x$.
By the exact same reasoning, we also have
\begin{align}
\left|q_t^x(x,a| x_{m+1}) - q_t(x,a| x_{m+1})\right|
&\leq \sum_{h=m+1}^{k(x)-1} \sum_{x_h', a_h', x_{h+1}'}
\epsilon_{i_t}^\star(x_{h+1}'|x_h', a_h') q_t(x_h', a_h' | x_{m+1}) q_t^x(x, a | x_{h+1}') \notag \\
&\leq \pi_t(a|x) \sum_{h=m+1}^{k(x)-1} \sum_{x_h', a_h', x_{h+1}'}
\epsilon_{i_t}^\star(x_{h+1}'|x_h', a_h') q_t(x_h', a_h' | x_{m+1})  \label{eqn:bound2}
\end{align}
Combining Eq.~\eqref{eqn:bound1} and Eq.~\eqref{eqn:bound2},
summing over all $t$ and $(x,a)$, and using the shorthands $w_m = (x_m, a_m, x_{m+1})$ and $w_h' = (x_h', a_h', x_{h+1}')$, we have derived
\begin{align}
&\sum_{t=1}^T \sum_{x\in X, a\in A} |q_t^x(x,a) - q_t(x,a)| \notag \\
&\leq \sum_{t,x,a}\sum_{m=0}^{k(x)-1}\sum_{w_m}
\epsilon_{i_t}^\star(x_{m+1}|x_m, a_m) q_t(x_m, a_m) q_t(x, a | x_{m+1}) \notag \\
&\hspace{2em} + \sum_{t,x,a}\sum_{m=0}^{k(x)-1}\sum_{w_m}
\epsilon_{i_t}^\star(x_{m+1}|x_m, a_m) q_t(x_m, a_m) \left(\pi_t(a|x) \sum_{h=m+1}^{k(x)-1} \sum_{w_h'}
\epsilon_{i_t}^\star(x_{h+1}'|x_h', a_h') q_t(x_h', a_h' | x_{m+1})\right)  \notag \\
&= \sum_{t}\sum_{k<L}\sum_{m=0}^{k-1}\sum_{w_m}
\epsilon_{i_t}^\star(x_{m+1}|x_m, a_m) q_t(x_m, a_m) \sum_{x \in X_k, a\in A}q_t(x, a | x_{m+1}) \notag \\
&\hspace{2em} + \sum_{t}\sum_{k<L}\sum_{m=0}^{k-1}\sum_{w_m}\sum_{h=m+1}^{k-1} \sum_{w_h'}
\epsilon_{i_t}^\star(x_{m+1}|x_m, a_m) q_t(x_m, a_m)
\epsilon_{i_t}^\star(x_{h+1}'|x_h', a_h') q_t(x_h', a_h' | x_{m+1})\left(\sum_{x \in X_k, a\in A}\pi_t(a|x)\right)   \notag \\
&= \sum_{0\leq m < k<L}\sum_{t, w_m}
\epsilon_{i_t}^\star(x_{m+1}|x_m, a_m) q_t(x_m, a_m) \notag \\
&\hspace{2em} + \sum_{0\leq m < h < k < L}|X_k|\sum_{t, w_m, w_h'}
\epsilon_{i_t}^\star(x_{m+1}|x_m, a_m) q_t(x_m, a_m)
\epsilon_{i_t}^\star(x_{h+1}'|x_h', a_h') q_t(x_h', a_h' | x_{m+1}) \notag \\
&\leq \underbrace{\sum_{0\leq m < k<L}\sum_{t, w_m}
\epsilon_{i_t}^\star(x_{m+1}|x_m, a_m) q_t(x_m, a_m)}_{\triangleq B_1} \notag \\
&\hspace{2em} + |X|\underbrace{\sum_{0\leq m < h < L}\sum_{t, w_m, w_h'}
\epsilon_{i_t}^\star(x_{m+1}|x_m, a_m) q_t(x_m, a_m)
\epsilon_{i_t}^\star(x_{h+1}'|x_h', a_h') q_t(x_h', a_h' | x_{m+1})}_{\triangleq B_2}. \notag
\end{align}
It remains to bound $B_1$ and $B_2$ using the definition of $\epsilon^\star_{i_t}$.
For $B_1$, we have
\begin{align*}
B_1 &= \scO\left(\sum_{0\leq m < k<L}\sum_{t, w_m}
q_t(x_m, a_m)\sqrt{\frac{P(x_{m+1}|x_m,a_m)\ln\left(\frac{T|X||A|}{\delta}\right)}{\max\{1,N_{i_t}(x_m,a_m)\}}} + \frac{q_t(x_m, a_m) \ln\left(\frac{T|X||A|}{\delta}\right)}{\max\{1,N_{i_t}(x_m,a_m)\}}
\right)  \\
&\leq \scO\left(\sum_{0\leq m < k<L}\sum_{t, x_m, a_m}
q_t(x_m, a_m)\sqrt{\frac{|X_{m+1}|\ln\left(\frac{T|X||A|}{\delta}\right)}{\max\{1,N_{i_t}(x_m,a_m)\}}} + \frac{q_t(x_m, a_m) \ln\left(\frac{T|X||A|}{\delta}\right)}{\max\{1,N_{i_t}(x_m,a_m)\}}
\right)  \\
&\leq \scO\left(\sum_{0\leq m < k<L} \sqrt{|X_m||X_{m+1}||A|T\ln\left(\frac{T|X||A|}{\delta}\right)} \right) \\
&\leq \scO\left(\sum_{0\leq m < k<L} \left(|X_m|+|X_{m+1}|\right)\sqrt{|A|T\ln\left(\frac{T|X||A|}{\delta}\right)} \right)  \\
&= \scO\left(L|X|\sqrt{|A|T\ln\left(\frac{T|X||A|}{\delta}\right)} \right),
\end{align*}
where the second line uses the Cauchy-Schwarz inequality, the third line uses Lemma~\ref{lem:aux}, and the fourth line uses the AM-GM inequality.

For $B_2$, plugging the definition of $\epsilon^\star_{i_t}$ and using trivial bounds (that is, $\epsilon^\star_{i_t}$ and $q_t$ are both at most $1$ regardless of the arguments), we obtain the following three terms (ignoring constants)
\begin{align*}
&\sum_{0\leq m < h < L}\sum_{t, w_m, w_h'}
\sqrt{\frac{P(x_{m+1}|x_m,a_m)\ln\left(\frac{T|X||A|}{\delta}\right)}{\max\{1,N_{i_t}(x_m,a_m)\}}} q_t(x_m, a_m)
\sqrt{\frac{P(x_{h+1}'|x_h',a_h')\ln\left(\frac{T|X||A|}{\delta}\right)}{\max\{1,N_{i_t}(x_h',a_h')\}}}  q_t(x_h', a_h' | x_{m+1}) \\
&+ \sum_{0\leq m < h < L}\sum_{t, w_m, w_h'}
\frac{q_t(x_m, a_m) \ln\left(\frac{T|X||A|}{\delta}\right)}{\max\{1,N_{i_t}(x_m,a_m)\}}
+ \sum_{0\leq m < h < L}\sum_{t, w_m, w_h'}
\frac{q_t(x_h', a_h') \ln\left(\frac{T|X||A|}{\delta}\right)}{\max\{1,N_{i_t}(x_h',a_h')\}}.
\end{align*}
The last two terms are both of order $\scO(\ln T)$ by Lemma~\ref{lem:aux} (ignoring dependence on other parameters),
while the first term can be written as $\ln\left(\frac{T|X||A|}{\delta}\right)$ multiplied by the following:
\begin{align*}
&\sum_{0\leq m < h < L}\sum_{t, w_m, w_h'}
\sqrt{\frac{q_t(x_m, a_m)P(x_{h+1}'|x_h',a_h') q_t(x_h', a_h' | x_{m+1})}{\max\{1,N_{i_t}(x_m,a_m)\}}}
\sqrt{\frac{q_t(x_m, a_m)P(x_{m+1}|x_m,a_m)q_t(x_h', a_h' | x_{m+1})}{\max\{1,N_{i_t}(x_h',a_h')\}}}  \\
&\leq  \sum_{0\leq m < h < L}\sqrt{\sum_{t, w_m, w_h'}\frac{q_t(x_m, a_m) P(x_{h+1}'|x_h',a_h')q_t(x_h', a_h' | x_{m+1})}{\max\{1,N_{i_t}(x_m,a_m)\}}}
\sqrt{\sum_{t, w_m, w_h'}\frac{q_t(x_m, a_m)P(x_{m+1}|x_m,a_m)q_t(x_h', a_h' | x_{m+1})}{\max\{1,N_{i_t}(x_h',a_h')\}}} \\
&= \sum_{0\leq m < h < L}\sqrt{|X_{m+1}|\sum_{t, x_m, a_m}\frac{q_t(x_m, a_m)}{\max\{1,N_{i_t}(x_m,a_m)\}}}
\sqrt{|X_{h+1}|\sum_{t, x_h', a_h'}\frac{q_t(x_h', a_h')}{\max\{1,N_{i_t}(x_h',a_h')\}}} \\
&= \scO\left(|A|\ln\left(\frac{T|X||A|}{\delta}\right)\right) \sum_{0\leq m < h < L} \sqrt{|X_m||X_{m+1}||X_h||X_{h+1}|} =
\scO\left(L^2|X|^2|A|\ln\left(\frac{T|X||A|}{\delta}\right)\right),
\end{align*}
where the second line uses the Cauchy-Schwarz inequality and the last line uses Lemma~\ref{lem:aux} again.
This shows that the entire term $B_2$ is of order $O(\ln T)$.
Finally, realizing that we have conditioned on the events stated in Lemmas~\ref{lem:confidence_width} and~\ref{lem:aux}, which happen with probability at least $1-6\delta$, finishes the proof.
\end{proof}

\subsection{Bounding $\regthree$ and $\regfour$}
\label{app:reg_three_four}

In this section, we complete the proof of our main theorem by bounding the terms $\regthree$ and $\regfour$.
We first state the following useful concentration lemma which is a variant of~\citep[Lemma~1]{neu2015explore} and is the key for analyzing the implicit exploration effect introduced by $\gamma$.
The proof is based on the same idea of the proof for~\citep[Lemma~1]{neu2015explore}.

\begin{lemma}\label{lem:IX}
For any sequence of functions $\alpha_{1}, \ldots, \alpha_T$ such that $\alpha_t \in [0, 2\gamma]^{X\times A}$ is $\calF_t$-measurable for all $t$, we have with probability at least $1 -\delta$,
\[
\sum_{t=1}^T\sum_{x,a} \alpha_t(x,a) \left(\ellhat_t(x,a) - \frac{q_t(x,a)}{u_t(x,a)}\ell_t(x,a)\right) \leq L\ln\tfrac{L}{\delta}.
\]
\end{lemma}
\begin{proof}
Fix any $t$.
For simplicity, let $\beta = 2\gamma$ and $\ind_{t,x,a}$ be a shorthand of $\ind\{x_{k(x)} = x, a_{k(x)} = a\}$.
Then for any state-action pair $(x,a)$, we have
\begin{equation}\label{eqn:ellhat_bound}
\ellhat_t(x,a) = \frac{\ell_t(x,a)\ind_{t,x,a}}{u_t(x,a)+\gamma}
\leq \frac{\ell_t(x,a)\ind_{t,x,a}}{u_t(x,a)+\gamma \ell_t(x,a)}
= \frac{\ind_{t,x,a}}{\beta}\cdot\frac{2\gamma \ell_t(x,a)/u_t(x,a)}{1+\gamma \ell_t(x,a)/u_t(x,a)} \leq \frac{1}{\beta}\ln\left(1 + \frac{\beta \ell_t(x,a)\ind_{t,x,a}}{u_t(x,a)}\right),
\end{equation}
where the last step uses the fact $\frac{z}{1+z/2} \leq \ln(1+z)$ for all $z\geq0$.
For each layer $k<L$, further define
\[
\Shat_{t,k} = \sum_{x\in X_k,a\in A} \alpha_t(x,a) \ellhat_t(x,a)
\qquad\text{and} \qquad
S_{t,k} = \sum_{x\in X_k,a\in A}\alpha_t(x,a) \frac{q_t(x,a)}{u_t(x,a)}\ell_t(x,a).
\]
The following calculation shows $\E_t\left[\exp(\Shat_{t,k})\right] \leq \exp(S_{t,k})$:
\begin{align*}
\E_t\left[\exp(\Shat_{t,k})\right]
&\leq \E_t\left[ \exp\left( \sum_{x\in X_k,a\in A} \frac{\alpha_t(x,a)}{\beta} \ln\left(1 + \frac{\beta \ell_t(x,a)\ind_{t,x,a}}{u_t(x,a)}\right)\right) \right]  \tag{by Eq.~\eqref{eqn:ellhat_bound}} \\
&\leq \E_t\left[ \prod_{x\in X_k,a\in A} \left(1 + \frac{\alpha_t(x,a)\ell_t(x,a)\ind_{t,x,a}}{u_t(x,a)}\right) \right] \\
&= \E_t\left[1 +  \sum_{x\in X_k,a\in A}\frac{\alpha_t(x,a)\ell_t(x,a)\ind_{t,x,a}}{u_t(x,a)}  \right] \\
&= 1 +  S_{t,k} \leq \exp(S_{t,k}).
\end{align*}
Here, the second inequality is due to the fact $z_1\ln(1+z_2) \leq \ln(1+z_1 z_2)$ for all $z_2 \geq -1$ and $z_1 \in [0,1]$, and we apply it with $z_1 = \frac{\alpha_t(x,a)}{\beta}$ which is in $[0,1]$ by the condition $\alpha_t(x,a)\in [0, 2\gamma]$;
the first equality holds since $\ind_{t,x,a} \ind_{t,x',a'} = 0$ for any $x \neq x'$ or $a \neq a'$ (as only one state-action pair can be visited in each layer for an episode).
Next we apply  Markov inequality and show
\begin{align}
\Pr\left[\sum_{t=1}^T (\Shat_{t,k} - S_{t,k}) > \ln\left(\frac{L}{\delta}\right) \right]
&\leq \frac{\delta}{L}\cdot \E\left[\exp\left(\sum_{t=1}^T (\Shat_{t,k} - S_{t,k})\right)\right]\notag \\
&=\frac{\delta}{L}\cdot\E\left[\exp\left(\sum_{t=1}^{T-1} (\Shat_{t,k} - S_{t,k})\right)\E_T\left[\exp\left(\Shat_{T,k} - S_{T,k}\right)\right] \right] \notag \\
&\leq \frac{\delta}{L}\cdot\E\left[\exp\left(\sum_{t=1}^{T-1} (\Shat_{t,k} - S_{t,k})\right) \right] \notag  \\
&\leq \cdots \leq \frac{\delta}{L}. 
\label{eqn:ellhat_bound_mid}
\end{align}
Finally, applying a union bound over $k=0, \ldots, L-1$ shows with probability at least $1-\delta$,
\[
\sum_{t=1}^T\sum_{x,a} \alpha_t(x,a) \left(\ellhat_t(x,a) - \frac{q_t(x,a)}{u_t(x,a)}\ell_t(x,a)\right) = \sum_{k=0}^{L-1} \sum_{t=1}^T (\Shat_{t,k} - S_{t,k}) \leq L\ln\left(\frac{L}{\delta}\right),
\]
which completes the proof.
\end{proof}

\paragraph{Bounding $\regthree$.}
To bound $\regthree=\sum_{t=1}^T \langle \qhat_t - q^*, \ellhat_t \rangle$, note that under the event of Lemma~\ref{lem:confidence_sets}, $q^* \in \cap_i \  \Delta(\calP_{i})$,
and thus $\regthree$ is controlled by the standard regret guarantee of OMD.
Specifically, we prove the following lemma.

\begin{lemma}\label{lem:OMD}
With probability at least $1-5\delta$, \alg ensures
$
\regthree = \scO\Big(\frac{L\ln(|X||A|)}{\eta} + \eta |X||A|T + \frac{\eta L\ln(L/\delta)}{\gamma}\Big).
$
\end{lemma}
\begin{proof}
By standard analysis (see Lemma~\ref{lem:Hedge} after this proof), OMD with KL-divergence ensures for any $q \in \cap_i \  \Delta(\calP_{i})$,
\[
\sum_{t=1}^T \langle \qhat_t - q, \ellhat_t \rangle \leq \frac{L\ln(|X|^2|A|)}{\eta} + \eta \sum_{t,x,a} \qhat_t(x,a)\ellhat_t(x,a)^2.
\]
Further note that $\qhat_t(x,a)\ellhat_t(x,a)^2$ is bounded by
\begin{align*}
\frac{\qhat_t(x,a)}{u_t(x,a) + \gamma}\ellhat_t(x,a) \leq \ellhat_t(x,a)
\end{align*}
by the fact $\qhat_t(x,a) \leq u_t(x,a)$.
Applying Lemma~\ref{lem:IX} with $\alpha_t(x,a) = 2\gamma$ then shows with probability at least $1-\delta$,
\[
\sum_{t,x,a} \qhat_t(x,a)\ellhat_t(x,a)^2 \leq \sum_{t,x,a} \frac{q_t(x,a)}{u_t(x,a)}\ell_t(x,a) + \frac{L\ln\tfrac{L}{\delta}}{2\gamma}.
\]
Finally, note that under the event of Lemma~\ref{lem:confidence_sets}, we have $q^* \in\cap_i \  \Delta(\calP_{i})$, $q_t(x,a) \leq u_t(x,a)$, and thus $\frac{q_t(x,a)}{u_t(x,a)}\ell_t(x,a) \leq 1$.
Applying a union bound then finishes the proof.
\end{proof}

\begin{lemma}\label{lem:Hedge}
The OMD update with $\qhat_1(x,a,x') = \frac{1}{|X_k||A||X_{k+1}|}$ for all $k<L$ and $(x,a,x') \in X_k \times A \times X_{k+1}$, and
\[\qhat_{t+1} = \argmin_{q\in\Delta(\calP_{i_t})}\; \eta \langle q, \ellhat_t \rangle + D(q \;\|\; \qhat_{t})\]
where $D(q \;\|\; q') =  \sum_{x,a,x'}q(x,a,x')\ln\frac{q(x,a,x')}{q'(x,a,x')}- \sum_{x,a,x'}\left(q(x,a,x')-q'(x,a,x')\right)$ ensures
\[
\sum_{t=1}^T \langle \qhat_t - q, \ellhat_t \rangle \leq \frac{L\ln(|X|^2|A|)}{\eta} + \eta \sum_{t,x,a} \qhat_t(x,a)\ellhat_t(x,a)^2
\]
for any $q \in \cap_i \  \Delta(\calP_{i})$, as long as $\ellhat_t(x,a) \geq 0$ for all $t,x,a$.
\end{lemma}

\begin{proof}
Define $\tilde{q}_{t+1}$ such that
\[
\tilde{q}_{t+1}(x, a, x') = \qhat_t(x,a,x') \exp\left(-\eta \ellhat_t(x,a)\right).
\]
It is straightforward to verify $\qhat_{t+1} = \argmin_{q\in\Delta(\calP_{i_t})} D(q \;\|\; \tilde{q}_{t+1})$ and also
\[
\eta\langle \qhat_t - q, \ellhat_t \rangle =  D(q \;\|\; \qhat_t) - D(q \;\|\; \tilde{q}_{t+1}) + D(\qhat_t \;\|\; \tilde{q}_{t+1}).
\]
By the condition $q \in \Delta(\calP_{i_t})$ and the generalized Pythagorean theorem we also have $D(q \;\|\; \qhat_{t+1})  \leq D(q \;\|\; \tilde{q}_{t+1})$ and thus
\begin{align*}
\eta\sum_{t=1}^T \langle \qhat_t - q, \ellhat_t \rangle
&\leq \sum_{t=1}^T \left( D(q \;\|\; \qhat_t) - D(q \;\|\; \qhat_{t+1}) + D(\qhat_t \;\|\; \tilde{q}_{t+1})\right) \\
&= D(q \;\|\; \qhat_1) - D(q \;\|\; \qhat_{T+1}) + \sum_{t=1}^T  D(\qhat_t \;\|\; \tilde{q}_{t+1}).
\end{align*}
The first two terms can be rewritten as
\begin{align*}
&\sum_{k=0}^{L-1}\sum_{x\in X_k}\sum_{a\in A}\sum_{x'\in X_{k+1}} q(x,a,x') \ln \frac{\qhat_{T+1}(x,a,x')}{\qhat_1(x,a,x')} \\
&\leq \sum_{k=0}^{L-1}\sum_{x\in X_k}\sum_{a\in A}\sum_{x'\in X_{k+1}} q(x,a,x') \ln (|X_k||A||X_{k+1}|) \tag{by definition of $\qhat_1$} \\
&=\sum_{k=0}^{L-1} \ln (|X_k||A||X_{k+1}|) \leq L\ln(|X|^2|A|).
\end{align*}
It remains to bound the term $D(\qhat_t \;\|\; \tilde{q}_{t+1})$:
\begin{align*}
D(\qhat_t \;\|\; \tilde{q}_{t+1})
&= \sum_{k=0}^{L-1}\sum_{x\in X_k}\sum_{a\in A}\sum_{x'\in X_{k+1}}
\left(\eta \qhat_t(x,a,x')\ellhat_t(x,a) - \qhat_t(x,a,x') + \qhat_t(x,a,x') \exp\left(-\eta \ellhat_t(x,a)\right)   \right) \\
&\leq \eta^2 \sum_{k=0}^{L-1}\sum_{x\in X_k}\sum_{a\in A}\sum_{x'\in X_{k+1}} \qhat_t(x,a,x')\ellhat_t(x,a)^2 \\
&= \eta^2 \sum_{x\in X, a\in A} \qhat_t(x,a)\ellhat_t(x,a)^2
\end{align*}
where the inequality is due to the fact $e^{-z} \leq 1- z + z^2$ for all $z \geq 0$.
This finishes the proof.
\end{proof}

\paragraph{Bounding $\regfour$.}
It remains to bound the term $\regfour=\sum_{t=1}^T \langle q^*, \ellhat_t - \ell_t \rangle$, which can be done via a direct application of Lemma~\ref{lem:IX}.

\begin{lemma}\label{lem:bias_2}
With probability at least $1-5\delta$, \alg ensures
$
\regfour = \scO\left(\frac{L\ln(|X||A|/\delta)}{\gamma}\right).
$
\end{lemma}
\begin{proof}
For each state-action pair $(x,a)$, we apply Eq.~\eqref{eqn:ellhat_bound_mid} in Lemma~\ref{lem:IX} with $\alpha_t(x',a') = 2\gamma \Ind{x'=x,a'=a}$, which shows that with probability at least $ 1 - \frac{\delta}{|X||A|}$,
\[
\sum_{t=1}^T \left(\ellhat_t(x,a) - \frac{q_t(x,a)}{u_t(x,a)}\ell_t(x,a)\right)\leq \frac{1}{2\gamma} \ln\left(\frac{|X||A|}{\delta}\right).
\]
Taking a union bound over all state-action pairs shows that with probability at least $1-\delta$, we have for all occupancy measure $q \in \Omega$,
\begin{align*}
\sum_{t=1}^T \left\langle q, \ellhat_t - \ell_t \right\rangle &\leq
\sum_{t,x,a} q(x,a)\ell_t(x,a)\left( \frac{q_t(x,a)}{u_t(x,a)} - 1 \right) + \sum_{x,a}\frac{q(x,a)\ln\tfrac{|X||A|}{\delta}}{2\gamma} \\
& = \sum_{t,x,a} q(x,a)\ell_t(x,a)\left( \frac{q_t(x,a)}{u_t(x,a)} - 1 \right) + \frac{L\ln\tfrac{|X||A|}{\delta}}{2\gamma}.
\end{align*}
%where $\regfour = \sum_{t=1}^T \left\langle q^\star, \ellhat_t - \ell_t \right\rangle$ is a special case when $q=q^\star$. 

Note again that under the event of Lemma~\ref{lem:confidence_sets}, we have $q_t(x,a) \leq u_t(x,a)$, so the first term of the bound above is nonpositive.
Applying a union bound and taking $q = q^\star$ finishes the proof.
\end{proof}

%%%%%%%%%%%%%%%%%%%%%%%%%%%%%%%%%%%%%%%%%%%%%%%%%%%%%%%%%%%%%%%%%%%%%%%%%%%%%%%
%%%%%%%%%%%%%%%%%%%%%%%%%%%%%%%%%%%%%%%%%%%%%%%%%%%%%%%%%%%%%%%%%%%%%%%%%%%%%%%
% DELETE THIS PART. DO NOT PLACE CONTENT AFTER THE REFERENCES!
%%%%%%%%%%%%%%%%%%%%%%%%%%%%%%%%%%%%%%%%%%%%%%%%%%%%%%%%%%%%%%%%%%%%%%%%%%%%%%%
%%%%%%%%%%%%%%%%%%%%%%%%%%%%%%%%%%%%%%%%%%%%%%%%%%%%%%%%%%%%%%%%%%%%%%%%%%%%%%%
%%%%%%%%%%%%%%%%%%%%%%%%%%%%%%%%%%%%%%%%%%%%%%%%%%%%%%%%%%%%%%%%%%%%%%%%%%%%%%%
%%%%%%%%%%%%%%%%%%%%%%%%%%%%%%%%%%%%%%%%%%%%%%%%%%%%%%%%%%%%%%%%%%%%%%%%%%%%%%%

\end{document}

\section{Proof of Theorem 2}
Fix the policy $\pi$, the target state-action pair $(\widehat{x},\widehat{a})$ and the confidence set $\mathbb{P}$ with counters $N$, $M$ and confidence interval $\epsilon$. Let $\tau$ be the index of the layer that $\widehat{x}$ belongs to, that is, $\tau = k(\widehat{x})$. With these notations, we then consider how to compute $u(\widehat{x},\widehat{a})$, the upper occupancy bound of $\mathbb{P}$ and $\pi$ at  $(\widehat{x},\widehat{a})$. 

Let $v^{P}(x)$ denote the conditional probability that $(x,a)$ is visited when state $x$ is visited through given $P$, that is $$ v^{P}(x) = \text{Pr}[ x_\tau =  x, a_\tau = a \mid x_{k(x)} = x; P,\pi], k(x) \leq \tau. $$ 
We use $v^*(x)$ to denote the maximized conditional probability $max_{P\in \mathbb{P}}v^P(x)$ within given confidence set $\mathbb{P}$.

Clearly, the upper occupancy bound of $(x,a)$ is $v^*(x_0)$. According to the loop-free property of MDPs, we only consider the layers before $X_\tau$. We propose the dynamic programming function $f$ as follows,
\begin{equation}
\label{eq:dynamic}
\begin{split}
f(x) = & \sum_{a\in A}\pi(a|x)\left[\max_{g_a} \sum_{x'\in X_{\tau+1}}g_a(x')f(x')\right],
\\
& \text{where} \ \ g_a\in [0,1]^{X_{\tau+1}}, \norm{g_a}_1=1, \\  \text{and} &\norm{g_a-\bar{P}(\cdot |x,a)}_1\leq \epsilon(x,a) \ \  \text{for} \ \  a \in A
\end{split}
\end{equation}
with the boundary condition $f(\widehat{x}) = \pi(\widehat{a}|\widehat{x})$. 

On the other hand, the $v^P(x)$ can be expanded one step further
\begin{equation}
\label{eq:belleq}
\begin{split}
v^{P}(x) & = Pr[ x_\tau =  x, a_\tau = a \mid x_{k(x)} = x] \\
& = \sum_{a\in A} \sum_{x' \in X_{k(x)+1}} 
\pi(a|x)P(x'|x,a)Pr[ x_\tau =  x, a_\tau = a \mid x_{k(x)+1} = x'] \\
& = \sum_{a\in A} \pi(a|x)\sum_{x' \in X_{k(x)+1}} 
P(x'|x,a) v^P(x')
\end{split}
\end{equation}
Similarly, the $v^*(x)$ can be formulated as follows, \begin{equation}
\label{eq:bellopt}
\begin{split}
v^*(x)  = \max_{P\in \mathbb{P}}\sum_{a\in A} \pi(a|x)\sum_{x' \in X_{k(x)+1}} 
P(x'|x,a) v^P(x')
\end{split}
\end{equation}

The idea of the dynamic programming approach is to decompose the overall maximization problem like Equation~\eqref{eq:bellopt} into several iterative maximization sub-problems as Equation~\eqref{eq:dynamic}. Our goal is to show that $v^*(x)$ equals to $f(x)$, in other words, it is legitimate to break the optimization problem in Equation~\eqref{eq:bellopt} into the problems stated in Equation~\eqref{eq:dynamic}. 
\vspace{15pt}

\textit{Proof}. First, we show that the constraint $P\in \mathbb{P}$ can be decomposed into several local constraints over local transition functions $P(\cdot|x,a)$ , and those constraints are independent. 

To prove this, we treat all entries $P(x'|x,a)$ as parameters where $(x,a,x')\in X_k \times A \times X_{k+1}$ for some $k\in0..\tau-1$. For each state-action pair $(x,a)$, we denote the local transition function $P(x'|x,a)$ as $\theta_{x,a}$, where $\theta_{x,a}$ is a distribution over $X_{k(x)+1}$. Similarly, denote the parameters of state $x$(or layer $k$) as $\theta_x$(or $\theta_k$). 

On the other hand, let $\Theta_{x,a}$(or $\Theta_x$, $\Theta_k$) be the set of all possible values of $\theta_{x,a}$(or $\theta_{x}$, $\theta_{k}$) by the constraint $P \in \mathbb{P}$. Especially, the $\Theta_{x,a}$ is defined as follows according to UCRL-2,
$$\Theta_{x,a} = \left\{ \theta \left| \theta \in Distribution(X_{k(x)+1}) \And \norm{\theta - \bar{P}(\cdot|x,a)}_1 \leq \epsilon(x,a) \right. \right\}$$
where the empirical transition function $\bar{P}$ and the confidence interval function $\epsilon$ are corresponding with confidence set $\mathbb{P}$.

To assemble a valid $P$ that belongs to $\mathbb{P}$, every $\theta_{x,a}$ should belong to $\Theta_{x,a}$, that is, 
$$ P \in \mathbb{P} \iff \forall (x,a)\in X\times A, P(x'|x,a) = \theta_{x,a} \in \Theta_{x,a}.$$
This equivalence guarantees the decomposition of original problem in Equation~\eqref{eq:dynamic} is legitimate in the way that the decision of transition function $P$ can be break into a group of decisions of local transition function $P(\cdot|x,a)$.

Then we decompose the optimization problem of $v^*(x)$ as follows 
\begin{equation*}
\begin{split}
v^*(x) & = \max_{ P \in \mathbb{P}} v^P(x) \\
 &= \max_{ \theta_0 \in \Theta_0,\cdots \theta_{\tau-1} \in \Theta_{\tau-1}} v^{\theta_{0:\tau-1}}(x) \text{ \ \ \ (decompose the decision of P)} \\ 
 & = \max_{ \theta_{k(x)} \in \Theta_{k(x)},\cdots \theta_{\tau-1} \in \Theta_{\tau-1}} v^{\theta_{k(x):K-1}}(x) 
 \text{ \ \ \ (remove layers prior to $k(x)$)} \\
 & = \max_{ \theta_x \in \Theta_x, \theta_{k(x)+1} \in \Theta_{k(x)+1},\cdots \theta_{\tau-1} \in \Theta_{\tau-1}} v^{\theta_x,\theta_{k(x):K-1}}(x)  \text{ \ \ \ (remove the decisions of other states)} \\
 & = \max_{ \theta_x \in \Theta_x, \theta_{k(x)+1} \in \Theta_{k(x)+1},\cdots \theta_{\tau-1} \in \Theta_{\tau-1}} \left[ \sum_{a\in A} \sum_{x' \in X_{k(x)+1}} 
\pi_(a|x)\theta_{x,a}(x') v^{\theta_{k(x)+1:K-1}}(x') \right] \\ 
& =  \max_{ \theta_x \in \Theta_x}\left[ \sum_{a\in A} \sum_{x' \in X_{k(x)+1}} 
\pi_(a|x)\theta_{x,a}(x') \max_{ \theta_{k(x)+1} \in \Theta_{k(x)+1},\cdots \theta_{\tau-1} \in \Theta_{\tau-1}} v^{\theta_{k(x)+1:K-1}}(x') \right] \\
& = \max_{ \theta_x \in \Theta_x}\left[ \sum_{a\in A} \sum_{x' \in X_{k(x)+1}} 
\pi_(a|x)\theta_{x,a}(x') v^*(x') \right] \\ & = \sum_{a\in A}\pi(a|x) \max_{ \theta_{x,a} \in \Theta_{x,a}}\left[ \sum_{x' \in X_{k(x)+1}} \theta_{x,a}(x') v^*(x') \right]
\end{split}
\end{equation*}
This equation indicate $v^*$ has optimal structure, that is, the maximization problem of $v^*(x)$ can be break into sub-maximization-problems for $v^*(x')$ in the next layers. It concludes that $v^*(x)$ equals to $f(x)$ for every state since the boundary condition $v^*(\widehat{x}) = \pi(\widehat{a}|\widehat{x})$ is the same.  

Therefore, this optimization problem can be solved by a backward dynamic programming process which computes the $v^*(x)$ from layer $\tau$ to layer $0$. At each iteration for $v^*(x)$ in layer $k$, it solves $|A|$ linear programming problems
\begin{equation*}
\begin{split}
&\max_{ \theta \in \Theta_{x,a}}\left[ \sum_{x' \in X_{k(x)+1}} \theta(x') v^*(x') \right] \\ 
\theta \in \Theta_{x,a} \Longleftrightarrow & \norm{\theta - \bar{P}(\cdot|x,a)}_1 \leq \epsilon(x,a), \norm{\theta}_1 = 1,\theta \in [0,1]^{X_{k(x)+1}}
\end{split}
\end{equation*}and update $v^*(x)$ with constant $\pi(a|x)$. This formulation is the same as Equation~\ref{eq:dynamic}.

In the rest of this section, we show that such linear programming problems can be solved in a greedy approach, and describe our solver in Algorithm~\ref{alg:greedy}.

\subsection{Greedy solver for linear programming}
The linear programming problem of state-action pair $(x,a)$ can be reformulated as follows
\begin{equation*}
\begin{split}
\textbf{Variables :}  \ & w_{x'}^{+}, w_{x'}^{-} \forall x' \in X_{k(x)+1} \\ 
\textbf{Constraints :} \ 
& w_{x'}^{+} \geq 0, w_{x'}^{-} \geq 0, \forall x' \in X_{k(x)+1} \\
& 0 \leq \bar{P}(x'|x,a) + w_{x'}^{+} - w_{x'}^{-} \leq 1, \forall x' \in X_{k(x)+1} \\
& \sum_{x'} \left(w_{x'}^{+}+w_{x'}^{-} \right)\leq \epsilon(x,a) \\
& \sum_{x'} \left[ \bar{P}(x'|x,a) + w_{x'}^{+} - w_{x'}^{-}\right] =1 + \sum_{x'} \left(w_{x'}^{+} - w_{x'}^{-}\right) =1 +  \sum_{x'} w_{x'}^{+} - \sum_{x'} w_{x'}^{-} =1 \\ 
\textbf{Objective :} \ & \sum_{x'} \left[ \bar{P}(x'|x,a) + w_{x'}^{+} - w_{x'}^{-}\right]v(x') =  \sum_{x'} \bar{P}(x'|x,a)v(x') + \sum_{x'} \left[w_{x'}^{+} - w_{x'}^{-}\right]v(x')
\end{split}
\end{equation*}
When $\epsilon(x,a)$ is small enough, the last two constraints actually imply that $\sum_{x'} w_{x'}^{+} = \sum_{x'} w_{x'}^{-} = \frac{1}{2}\epsilon(x,a)$. Then we can use a greedy algorithm to solve this linear programming problem : first assign $w_{x'}^{+}$ as large as it allows in the descending order of $v(x')$, and then assign $w_{x'}^{-}$ in the ascending order of $v(x')$. 

We briefly explain why this greedy algorithm achieves optimal solution: suppose the optimal assignment $w$ is better from the computed $\hat{w}$, there must be $x'$ that is the first state in the descending order of $v(x')$ satisfies $w^+_{x'} > \hat{w}^+_{x'} $. Consider the sum of all the $w^+_{x'}$ which priors to $x$, it must be equal to $\frac{1}{2}\epsilon(x,a)$ according to the greedy algorithm. By contradicting with the constraint, the optimality of $\hat{w}$ is proved.

\newpage

\begin{algorithm}[H]
\label{alg:greedy} 
\textbf{Input:} state $x$, action $a$, time horizon T, confidence parameter $\delta$, estimate transition function $\bar{P}$, epsilon function $\epsilon$, policy $\pi$, state-action counter $N$, state-action-afterstate counter $M$

\textbf{Initialization:} \\
\Indp $ K = k(x)$ \; 
initialize value function $\forall x\in X: \ v(x) = 0$\;
initialize target policy $\forall (x,a,x')$: $P_{x,a}(x'|x,a) \gets 0$ \;
set up boundary condition : $v(x)=1$

\Indm\For{ $k = K-1 \ \textbf{to} \ 0$ }{
\For{ $(x,a) \in X_k \times A $ }{
\eIf{$\epsilon(x,a) \leq 1$}{
  initialize temporary weight vector $ \forall x' \in X_{k+1}$: $w(x') = \bar{P}(x,a,x') $ \;
  initialize temporary weight counter: $ \Delta_{up} = \frac{1}{2}\epsilon(x,a)$ \;
  
  \For{ $x' \in X_{k+1}$ in \textbf{descending} order of $v(x')$}{
    \eIf{$\Delta_{up} \neq 0$ }{
    $ \Delta_{x'} = \min\{ \Delta_{up}, 1 - w(x')\}$ \;
    $ w(x') \gets w(x') + \Delta_{x'}$ \;
    $ \Delta_{up} \gets \Delta_{up} - \Delta_{x'}$
    }{ stop looping \;
    }}
    
    $\Delta_{down} = \frac{1}{2}\epsilon(x,a) $ \;
    
    \For{ $x' \in X_{k+1}$ in \textbf{ascending} order of $v(x')$}{
    \eIf{$\Delta_{down} \neq 0$ }{
    $ \Delta_{x'} = \min\{ \Delta_{down}, w(x')\}$ \;
    $ w(x') \gets w(x') - \Delta_{x'}$ \;
    $ \Delta_{down} \gets \Delta_{down} - \Delta_{x'}$
    }{ stop looping \;
    }}
    
    $v(x) \gets v(x) + \pi(a|x)\sum_{x'\in X_{k+}}w(x')v(x')$ \;
    update target policy $\forall x' \in X_{k+1}$: $P_{x,a}(x'|x,a) \gets w(x')$ \;
  }{
    $v(x) \gets v(x) + \pi(a|x) \max_{x' \in X_{k_1}}v(x')$ 
  } }}
\textbf{Output:} return $v(x_0)$ and $P_{x,a}$ \\
 \caption{Comp-Prob-UB Algorithm}
\end{algorithm}

\section{Proof of Theorem 6}
\begin{lemma}
For any $ 0 < \delta < 1$
\begin{equation}
\label{eq:confidencesets}
\norm{ P(\cdot|x,a) - \bar{P_i}(\cdot|x,a) }_1 \leq \sqrt{\frac{2\left|X_{k(x)+1}\right| \ln{\frac{T|X||A|}{\delta}}}{\max\{1,N_i(x,a)\}}} 
\end{equation}
holds with probability at least $ 1 - \delta $ simultaneously for all $(x,a) \in X \times A$ and all epochs.
\end{lemma} 

In addition, the previous paper of UC-O-REPS proves the following theorem, which further ensures that the estimation error caused by transition estimation can be bounded with $\order(L|X|\sqrt{|A|T})$. 

\begin{lemma}
Denote $\mathbb{P}_t$ be the confidence set of transition probabilities induced by $\Delta(M,i(t))$. 
Let $\{ \pi_t \}_{t=1}^{T}$ be policies and let $\{ P_t \}_{t=1}^{T}$ be transition functions such that $P_t \in \mathbb{P}_t$ for every t. Then with probability $1 - 2\delta$,
$$
\sum_{t=1}^{T} \norm{ q^{P_t,\pi_t} - q^{P,\pi_t} }_1 \leq 3 L|X|\sqrt{2T|A|\ln{\frac{T|X||A|}{\delta}}} + 2L|X|\sqrt{2T\ln{\frac{L}{\delta}}} 
$$
\end{lemma}
Proof. Lemma B.2 and Lemma B.3 in Appendix of "Online Convex Optimization in Adversarial Markov Decision Processes". Though the algorithm is changed, the theorem still holds since its proof doesn't require any constraints on $\{P_t\}_{t=1}^{T}$ other than Condition~\eqref{eq:confidencesets}.

\section{Regret Analysis}

\section{Linear Constraints}
\begin{equation*}
\begin{split}
 \sum_{x\in X_k}\sum_{a\in A}\sum_{x'\in X_{k+1}}q(x,a,x') = 1  &\forall k = 0, \cdots, L-1 \\
 \sum_{x'\in X_{k-1}}\sum_{a\in A}q(x',a,x) =  \sum_{a\in A}\sum_{x'\in X_{k+1}}q(x,a,x')   & \forall x \in X_k \\
 q(x,a,x') - \overline{P_i}(x'|x,a)\sum_{y\in X_{k+1}}a(x,a,y) \leq \epsilon(x,a,x') &  \forall(x,a,x')\in X_k \times A \times X_{k+1} \\ 
 \overline{P_i}(x'|x,a)\sum_{y\in X_{k+1}}a(x,a,y) - q(x,a,x') \leq \epsilon(x,a,x')& \forall(x,a,x')\in X_k \times A \times X_{k+1} \\ 
 \sum_{x'\inX_{k+1}}\epsilon(x,a,x') \leq \epsilon_i(x,a) \sum_{x'\in X_{k+1}}q(x,a,x') & \forall(x,a)\in X_k \times A \\
 q(x,a,x') \geq 0 & \forall(x,a,x')\in X_k \times A \times X_{k+1} 
\end{split}
\end{equation*}